%% file: 0_main.tex
\documentclass{ieeeaccess}
\usepackage{cite}
\usepackage{amsmath,amssymb,amsfonts}
\usepackage{algorithmic}
\usepackage{graphicx}
\usepackage{textcomp}
\usepackage{bm} 
\usepackage{caption}
\usepackage{soul,xcolor}
\usepackage{microtype}
\usepackage{booktabs}
\usepackage{siunitx}
\usepackage{hyperref}
\usepackage{acro}
\newcommand{\review}[1]{\textcolor{black}{#1}}
\input{include/acronyms.tex}

\def\BibTeX{{\rm B\kern-.05em{\sc i\kern-.025em b}\kern-.08em
    T\kern-.1667em\lower.7ex\hbox{E}\kern-.125emX}}
\begin{document}

\history{Date of publication 04 March 2024, date of current version 24 March 2024.}
\doi{\href{https://doi.org/10.1109/ACCESS.2024.3373004}{10.1109/ACCESS.2024.3373004}}

\title{ ForestTrav: 3D LiDAR-only Forest Traversability Estimation for Autonomous Ground Vehicles}
\author{
    \uppercase{Fabio Ruetz}\authorrefmark{1,2},
    \uppercase{Nicholas Lawrance}\authorrefmark{2}, \IEEEmembership{Member, IEEE},
    \uppercase{Emili Hern\'andez}\authorrefmark{3}, \IEEEmembership{Member, IEEE}, 
    \uppercase{Paulo Borges}\authorrefmark{2}, 
    and \uppercase{Thierry Peynot}\authorrefmark{1}, \IEEEmembership{Senior Member, IEEE}
}
\address[1]{ QUT Centre for Robotics, Queensland University of Technology (QUT), Brisbane Qld 4000, Australia.}
\address[2]{ CSIRO Robotics, Data61, Pullenvale, Qld 4069, Australia.}
\address[3]{ Emesent, Milton, Qld 4064, Australia.}
\tfootnote{This work was supported by QUT, CSIRO, Emesent and the SILVANUS Project through European Commission Funding on the Horizon 2020 call number H2020-LC-GD-2020, Grant Agreement number 101037247. F.R. and T.P. acknowledge continued support from the Queensland University of Technology (QUT) through the Centre for Robotics.}

\markboth
{Author \headeretal: Preparation of Papers for IEEE TRANSACTIONS and JOURNALS}
{Author \headeretal: Preparation of Papers for IEEE TRANSACTIONS and JOURNALS}

\corresp{Corresponding author: Fabio Ruetz (e-mail:fabioadrian.ruetz@qut.edu.au).}

\begin{abstract}
Autonomous navigation in unstructured vegetated environments remains an open challenge. To successfully operate in these settings, autonomous ground vehicles (AGVs) must assess the environment and determine which vegetation is pliable enough to safely traverse. In this paper, we propose ForestTrav
(\textbf{Forest Trav}ersability) : a novel lidar-only (geometric), online traversability estimation (TE) method that can accurately generate a per-voxel traversability estimate for densely vegetated environments, demonstrated in dense subtropical forests. The method leverages a salient, probabilistic 3D voxel representation, continuously fusing incoming lidar measurements to maintain multiple, per-voxel ray statistics, in combination with the structural context and compactness of sparse convolutional neural networks (SCNNs) to perform accurate TE in densely vegetated environments. The proposed method is real-time capable and is shown to outperform state-of-the-art volumetric and 2.5D TE methods by a significant margin (0.62 vs. 0.41 Matthews correlation coefficient (MCC) score at \qty{0.1}{\m} voxel resolution) in challenging scenes and to generalize to unseen environments. ForestTrav demonstrates that lidar-only (geometric) methods can provide accurate, online TE in complex, densely-vegetated environments. This capability has not been previously demonstrated in the literature in such complex environments. 
Further, we analyze the response of the TE methods to the temporal and spatial evolution of the probabilistic map as a function of information accumulated over time during scene exploration. It shows that our method performs well even with limited information in the early stages of exploration, and this provides an additional tool to assess the expected performance during deployment.
Finally, to train and assess TE methods in highly-vegetated environments, we collected and labeled a novel, real-world data set and provide it to the community as an open-source resource. 

\end{abstract}


\titlepgskip=-15pt

\maketitle
\input{01_introduction.tex}

\input{02_related_work.tex}

\input{03_methodology.tex}

\input{04_experiments_results.tex}

\input{05_discussion_conclusion.tex}


\addtolength{\textheight}{-0cm}   

\bibliographystyle{IEEEtran}
\bibliography{references}


\begin{IEEEbiography}[{\includegraphics[width=1in,height=1.25in,clip,keepaspectratio]{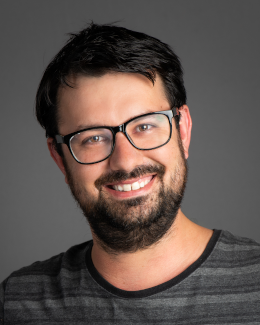}}]{Fabio A. Ruetz} 
received his B.S and M.S degrees in Mechanical and Process Engineering from Swiss Federal Institute of Technology Zurich (ETH), Zurich, Switzerland, in 2018. Since May 2020, he is pursuing a Ph.D. at QUT Centre for Robotics, Queensland University of Technology (QUT), in collaboration with Robotics and Autonomous Systems Group (RASG), CSIRO, and Emesent. His research and interest lie in probabilistic mapping, computer vision, machine learning, and path planning to enable autonomous ground vehicles to operate in challenging environments. 
\end{IEEEbiography}

\begin{IEEEbiography}[{\includegraphics[width=1in,height=1.25in,clip,keepaspectratio]{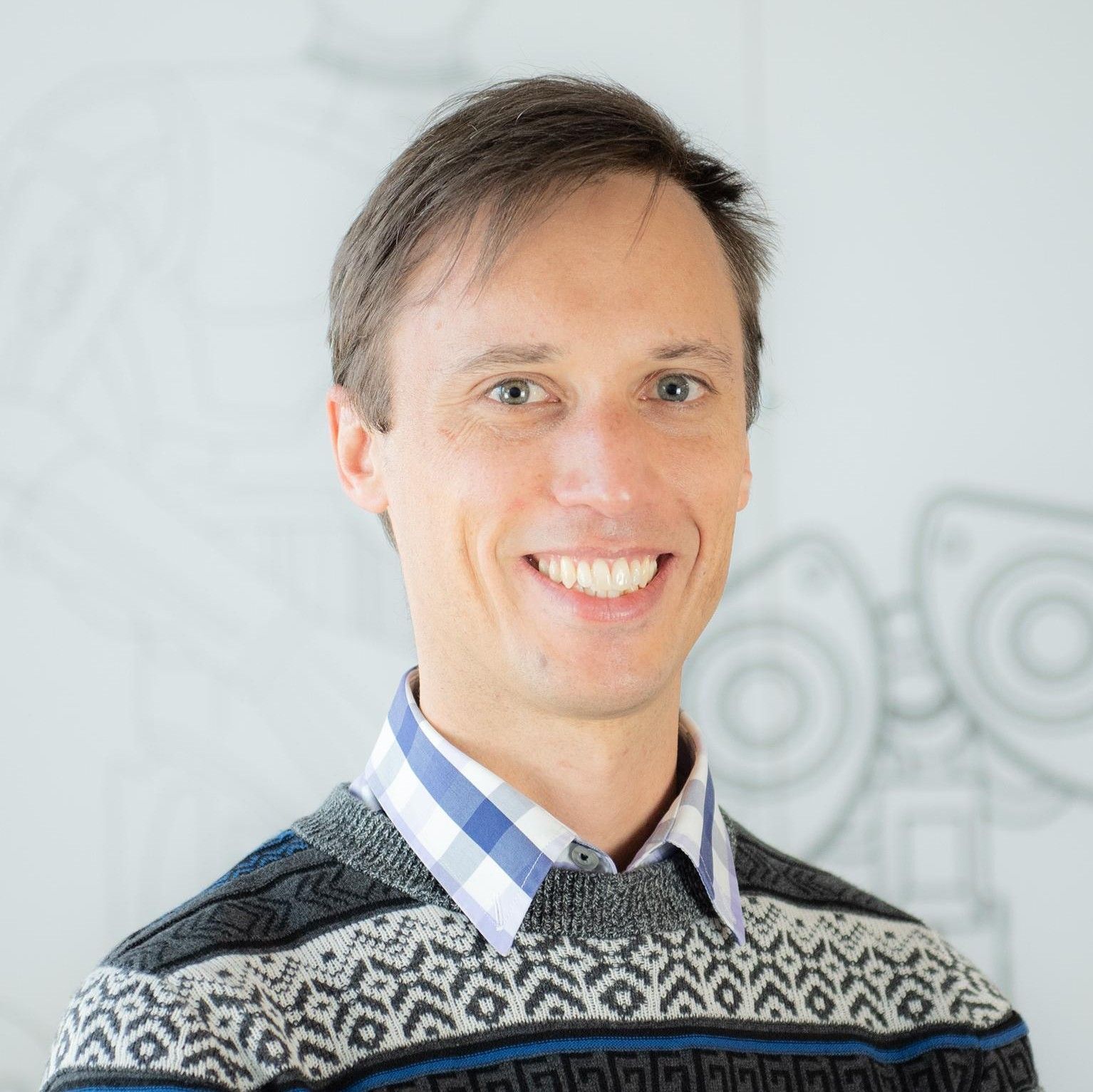}}]{Nicholas Lawrance}
completed his PhD at the University of Sydney and worked as a postdoctoral scholar at Oregon State University, USA and ETH Zurich, Switzerland. He is currently a senior research scientist in robotic perception and autonomy at the Commonwealth Scientific and Industrial Research Organisation (CSIRO) in Australia. His research focuses on adaptive planning approaches for mobile robots, particularly in the presence of environmental uncertainty. Research interests include stochastic reasoning, adaptive sampling, and modelling of complex, uncertain phenomena. Applications include aerial, ground and underwater domains, particularly for long-duration robotic missions. Nick is a member of IEEE, an Associate Editor of IEEE Robotics and Automation Letters (RA-L), and a former Associate Editor for the International Conference on Robotics and Automation (ICRA).
\end{IEEEbiography}

\begin{IEEEbiography}[{\includegraphics[width=1in,height=1.25in,clip,keepaspectratio]{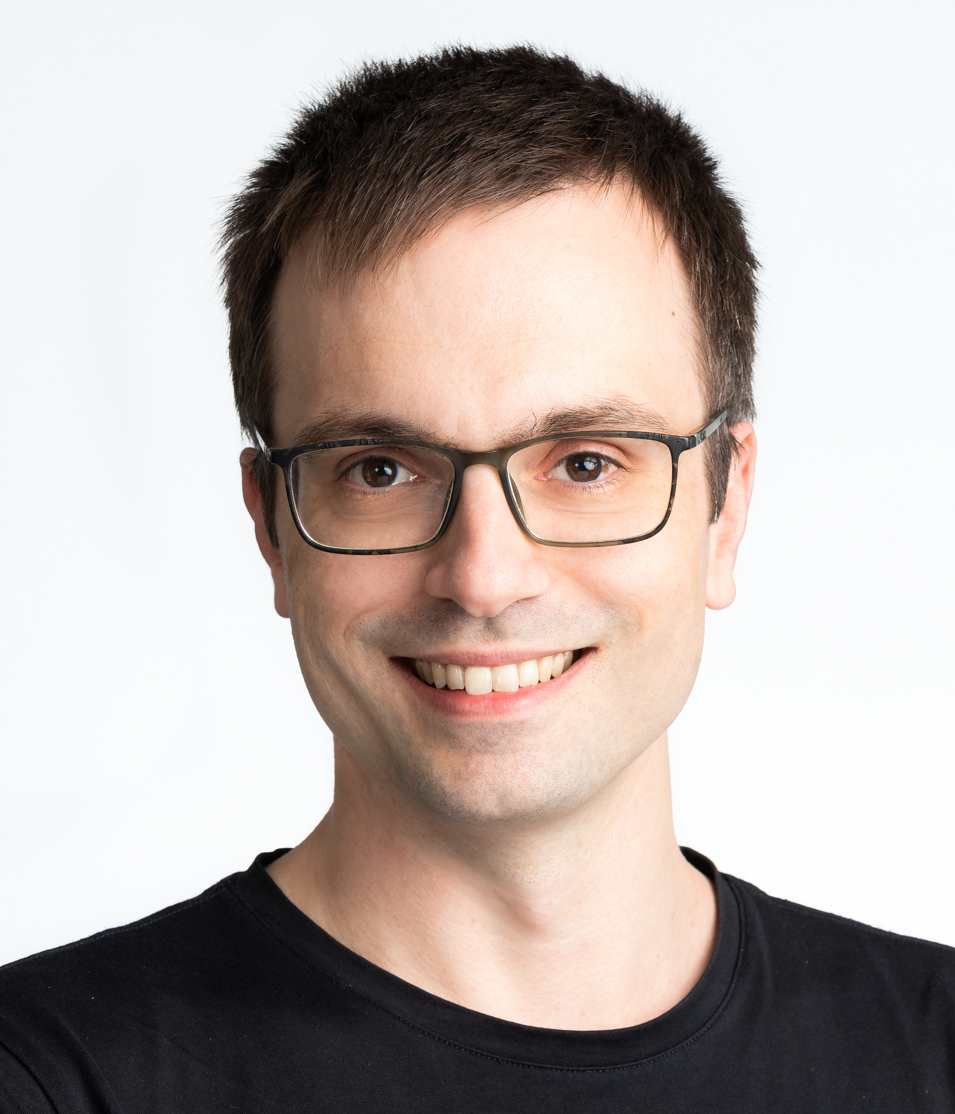}}] {Emili Hern\'andez} is an R\&D Manager with Emesent. He has two decades of experience on designing, developing and deploying novel software algorithms for underwater, ground and aerial robots. His current focus is on commercializing robotic autonomy research outcomes to improve and automate data capture in underground mining and asset inspection operations. He got his PhD at the University of Girona, Spain, and worked in several research positions at the CSIRO's Robotics and Autonomous Systems Group, Australia.
\end{IEEEbiography}

\begin{IEEEbiography}[{\includegraphics[width=1in,height=1.25in,clip,keepaspectratio]{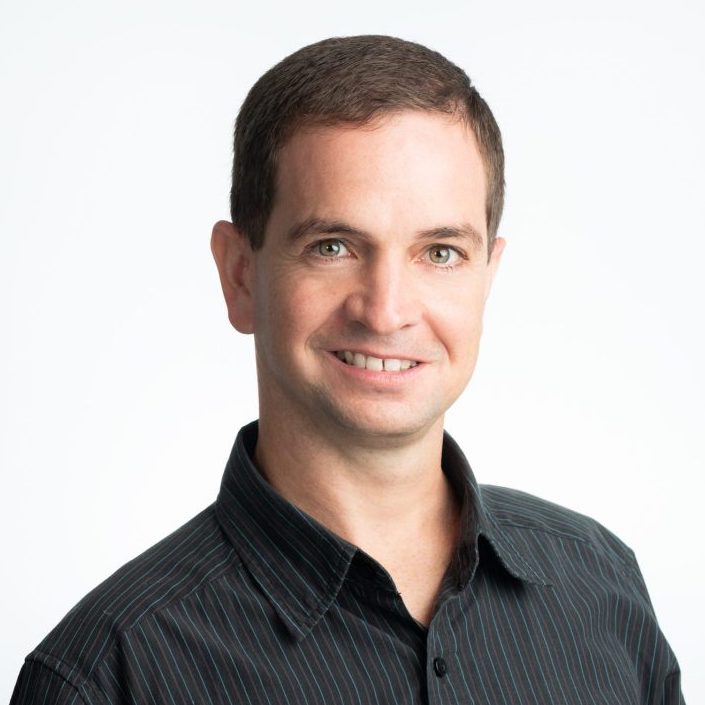}}]{Paulo V. K. Borges}
    is a Principal Research Scientist in the Robotics and Autonomous Systems Group at CSIRO (Brisbane, Australia). He has a Ph.D. in Electronic Engineering from Queen Mary, University of London (2007). Paulo has lived and worked in different countries (USA, Brazil, UK, Switzerland, and Australia), with work experience at CSIRO, University of London, Federal University of Santa Catarina, and  University of Manchester. He has also held visiting scientist positions at NASA Ames (2022-23) and ETH Zurich (2012-13). He currently holds adjunct positions as an Associate Professor at the University of Queensland and at Griffith University. In the last several years his core interest has been in autonomous robot solutions for the manufacturing, energy, and agriculture industries, with close connections between industry and research. 
\end{IEEEbiography}

\begin{IEEEbiography}[{\includegraphics[width=1in,height=1.25in,clip,keepaspectratio]{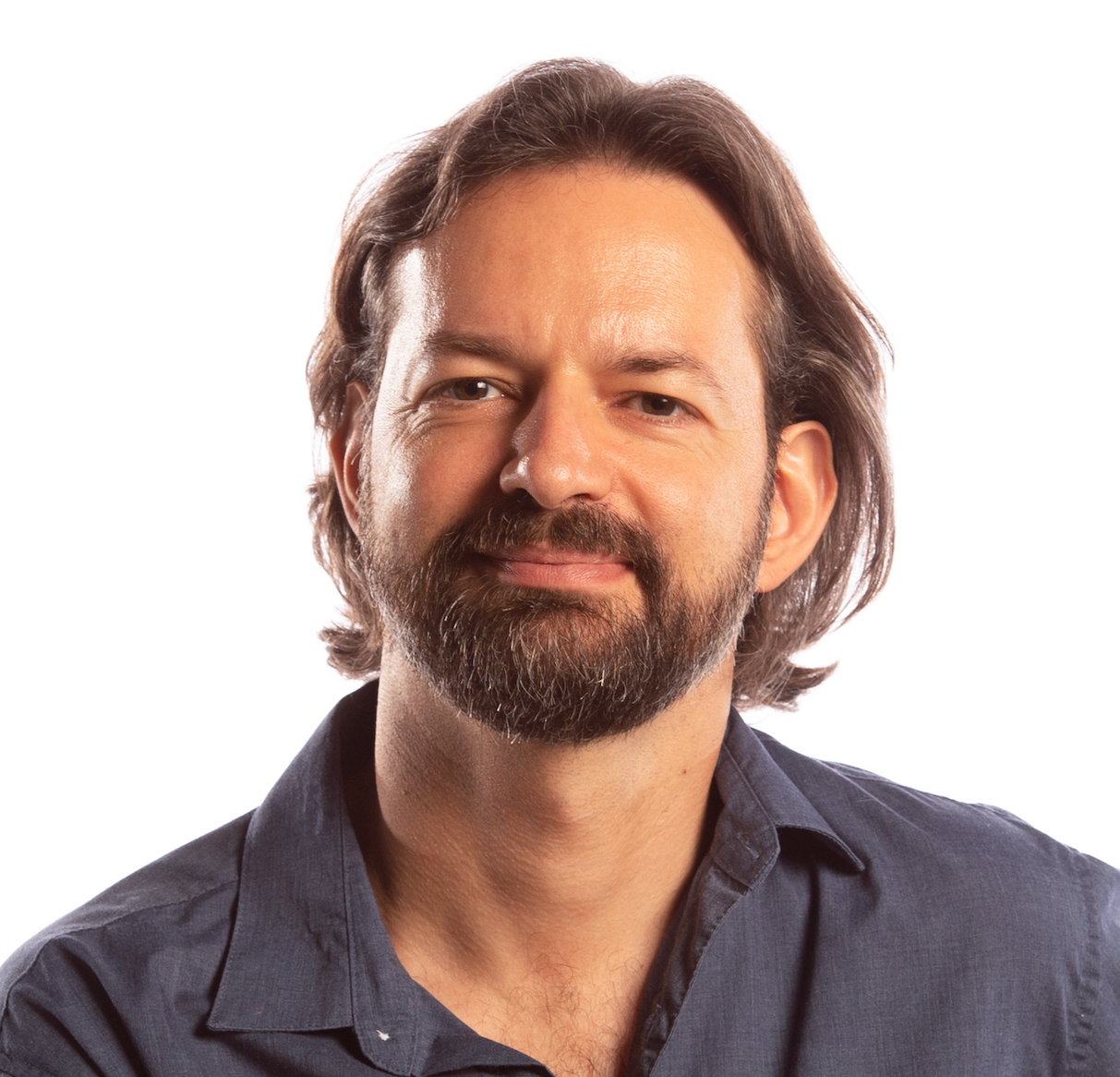}}]{Thierry Peynot}
    obtained his Ph.D. from the University of Toulouse and LAAS-CNRS in France. 
    He is Associate Professor in Robotics and Autonomous Systems at Queensland University of Technology (QUT) and a Chief Investigator of the QUT Centre for Robotics, where he leads the Mining Robotics and Space Robotics activities. Prior to joining QUT he was a researcher at the Australian Centre for Field Robotics (ACFR), The University of Sydney, worked at NASA Ames. Thierry has led multiple research programs funded by government, research institutions and industry, including mining (e.g. Caterpillar, Komatsu, Mining3), defence (e.g. BAE Systems, Rheinmetall) and space (e.g. with Boeing and CSIRO), developing robust perception technology for field robots and autonomous vehicles that can function despite adverse environmental conditions. 
    Thierry is a senior member of IEEE, immediate past Chair of the Robotics and Automation / Control Systems chapter, IEEE Queensland Section, and is a former Associate Editor of IEEE Robotics and Automation Letters (RA-L), the International Conference on Robotics and Automation (ICRA) and the International Conference on Intelligent Robots and Systems (IROS) . 
\end{IEEEbiography}

\EOD

\end{document}

%% file: include/acronyms.tex
\DeclareAcronym{SVM}{
  short = SVM,
  long  = support vector machine,
  short-indefinite = an,
  long-indefinite = a
}

\DeclareAcronym{AGV}{
  short = AGV,
  long  =  autonomous ground vehicle,
  short-indefinite = an,
  long-indefinite = a
}

\DeclareAcronym{SCNN}{
  short = SCNN,
  long  = sparse convolutional neural network,
  short-indefinite = an,
  long-indefinite = a
}

\DeclareAcronym{NDT-OM}{
  short = NDT-OM,
  long  = normal distributions transform occupancy maps,
    short-indefinite = an,
  long-indefinite = a  
}

\DeclareAcronym{NDT-TM}{
  short = NDT-TM,
  long  = normal distributions transform traversability maps,
  short-indefinite = an,
  long-indefinite = a  
}
\DeclareAcronym{NDT}{
  short = NDT,
  long  = normal distributions transform,
  short-indefinite = an,
  long-indefinite = a    
}

\DeclareAcronym{OHM}{
  short = OHM,
  long  = occupancy homogeneous map,
  short-indefinite = an,
  long-indefinite = an  
}

\DeclareAcronym{FTM}{
  short = FTM,
  long  = normal distributed transform forest traversability mapping,
  short-indefinite = an,
  long-indefinite = a  
}

\DeclareAcronym{TE}{
  short = TE,
  long  = traversability estimation,
  short-indefinite = a,
  long-indefinite = a  
}

\DeclareAcronym{MCC}{
  short = MCC,
  long  = Matthews correlation coefficient,
  short-indefinite = an,
  long-indefinite = a  
}

\DeclareAcronym{lfe}{
  short = LfE,
  long  = learning from experience,
  short-indefinite = an,
  long-indefinite = a  
}
\DeclareAcronym{vd}{
  short = CVD,
  long  = column vegetation density,
  short-indefinite = a,
  long-indefinite = a  
}


%% file: 01_introduction.tex
\section{Introduction}
\label{sec_introduction}

Autonomous navigation of ground vehicles in unstructured vegetated environments is essential for many robotics applications but remains an open challenge. Any off-road autonomous navigation requires accurate \acf{TE}, which can be defined as the determination of which part of the environment a given robot can or cannot travel through safely.  
Existing approaches often assume the environment is rigid. However, in natural environments, this assumption is overly prohibitive as successful navigation requires the robot to push and pass through pliable vegetation. 

\begin{figure}
    \centering
    \includegraphics[width=\columnwidth]{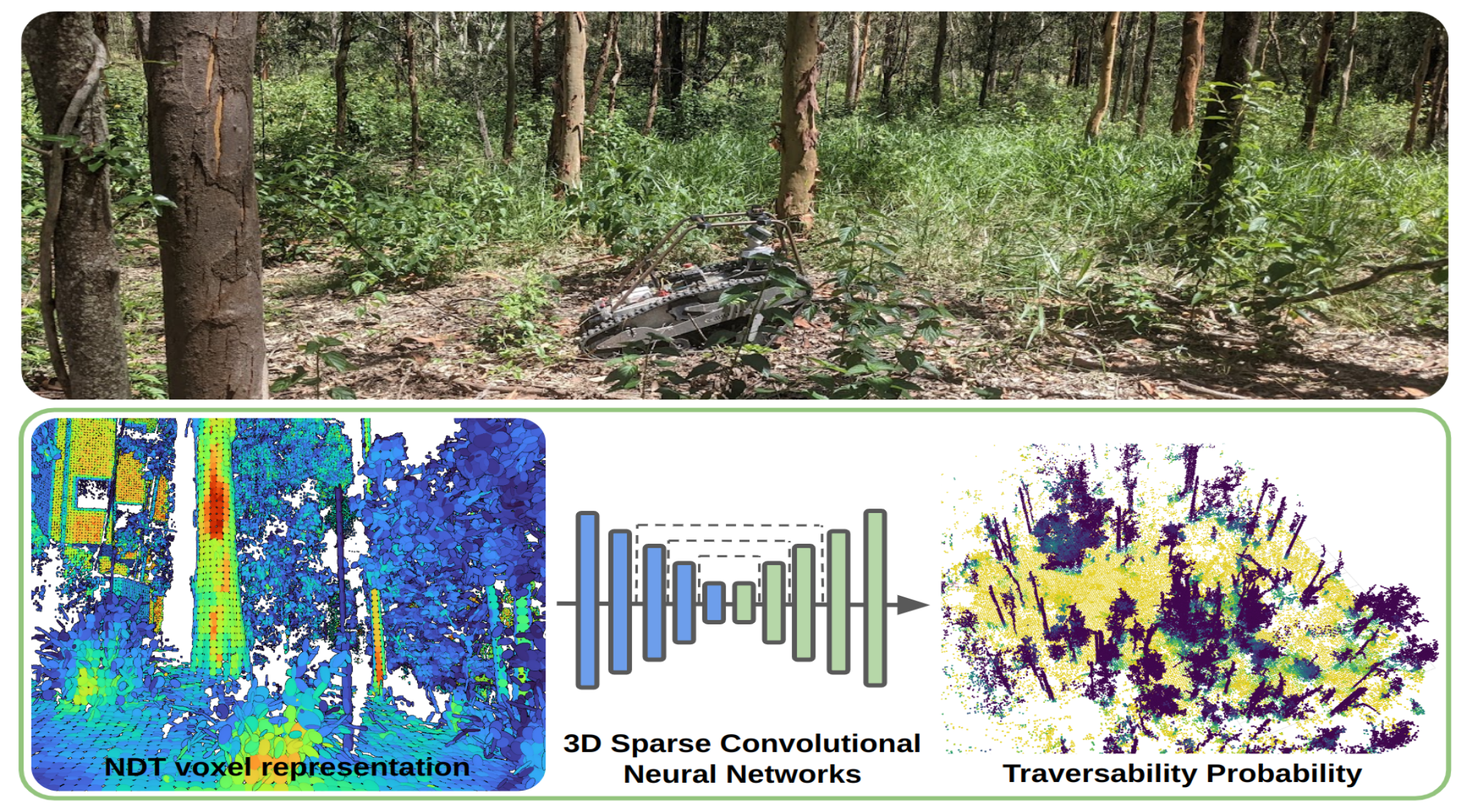}
    \caption{Top: robot navigating in a forest with low to dense vegetation. Bottom: traversability estimation using ForestTrav. Lidar data are accumulated into a probabilistic 3D voxel map. The map is then passed to a sparse convolutional neural network to estimate per-voxel traversability.} 
    \label{fig:intro_img}
\end{figure}
Unstructured vegetated environments such as forests are also challenging to navigate through because vegetation clutter can result in complex geometric structures with overhanging elements and occlusions. \ac{TE} in such environments requires a high-fidelity representation of the environment and a method that can leverage the representation with computational efficiency and generate accurate predictions. 

Navigation strategies in unstructured environments without vegetation, relying on high-fidelity geometric representation, have proven sufficient in various complex scenarios~\cite{chung2023into}. These approaches commonly represent the environment using 2D or 2.5D maps (one elevation value $z$ per position $(x,y)$ on a plane), grid maps (discretized 2D map), or elevation maps to assess traversability~\cite{HudTal21}. These methods rely on geometric features or heuristics generated from these representations. However, 2.5D maps cannot represent complex vegetated environments sufficiently. Thus, geometric \ac{TE} approaches have often been considered insufficient in the recent literature, and image-based modalities have been considered essential for accurate \ac{TE}~\cite{kahn2020badgr, ahtiainen2017normal, bradley2015scene, wellhausen2019where}. 

Single-image appearance-based TE methods are adequate for many tasks, e.g. following gravel roads with sparse vegetation and navigating greenhouses~\cite{schmid2022self, uzawa2022end}. They can leverage high-fidelity information contained in images in conjunction with strong assumptions about the environments~\cite{kahn2020badgr, wellhausen2019where}. However, these methods can only assess the next step locally and discard previous information. Hence, single image-based methods are more prone to failure than probabilistic map representations, due to sensor occlusions or image degradation~\cite{bradley2015scene}. 

Hybrid methods aim to combine geometric and appearance-based methods, leveraging the discriminative power of image-based modalities with geometric representations~\cite{bradley2015scene, frey2023fast, li2023seeing, maturana2018real}. However, the underlying geometric component fails to adequately represent the environments, degrading performance of the image-based components, as shown in Fig.~\ref{fig:2d_como_qulaitative} and the accompanying video.

In previous work~\cite{ruetz2022FTM}, we demonstrated the feasibility of \ac{TE} in complex vegetated environments using a 3D voxel-based representation containing multiple ray statistics, color fusion and consideration of a voxel's neighborhood. While the method showed strong results in offline processing, it was unsuitable for real-world deployment due to computational cost constraints and performance degradation while exploring a new environment. The performance of the method also showed limitations in the highly challenging, densely vegetated environments such as those addressed in this work. 
\acp{SCNN} have shown low inference times for sparse voxel representations, which are suitable for \ac{TE} estimation and scene completion in structured environments~\cite{frey2022locomotion}. However, their use has not been explored for \ac{TE} in vegetated environments. Our work leverages these findings and presents a novel \ac{TE} approach suitable for real-world deployment (in terms of accuracy and inference time) in complex vegetated environments such as those shown in Fig.~\ref{fig:intro_img}. 

\review{In this work, we claim the following contributions.}
\begin{itemize}
    \item \review{We propose a novel method for traversability estimation named ForestTrav (\textbf{Forest Trav}ersability). The proposed method represents the environment with a high-fidelity and feature-rich 3D voxel representation relying only on range sensor data. In addition, it leverages structural context and sparseness properties of \acp{SCNN} to allow for rapid inference. The small model size (approximately 2 million parameters [\qty{1.7}{\mega\byte}] per model) allows for data- and time-efficient training and deployment. The data set size required for training is economical and the method is trained exclusively on real-world data.}
    
    \item \review{We demonstrate the suitability of the proposed method for TE in complex natural terrain such as a highly vegetated forest environment. The proposed method is thoroughly evaluated on a challenging real-world data set and is shown to significantly outperform state-of-the-art (SOTA) methods in complex scenes and to generalize to unseen environments.}
    
    \item \review{We propose an evaluation metric to assess the robustness of an estimator's performance with respect to the temporally evolving, 3D probabilistic map. It quantifies performance changes that can occur while the robot moves into a new, unseen area. This is not addressed in the current \ac{TE} literature.}
    
    \item \review{Finally, we provide open-source access to our novel data set containing voxelized input features and output labels generated by our accurate labeling method. We provide nine different forest environments with varying degrees of vegetation. Our labeling method combines robot experience and expert domain knowledge in a probabilistic fashion.}
\end{itemize}

\review{This work and contributions address the research gap of \ac{TE} for \acp{AGV} in vegetated environments, specifically in full 3D leveraging only range sensors. The presented method can be considered a ``geometric'' method, relying solely on lidar measurements. Most prior works claim that \ac{TE} in vegetated environments requires appearance-based modalities. We show that this is not necessarily true, and our approach provides a highly accurate geometric solution shown to outperform SOTA methods. We believe that making the data available is extremely important as we have observed that the definition and complexity of ``forest environments'' varies significantly across different works in the literature. Although subjective, we argue that the forest navigation scenarios dealt with in this paper,  which can be seen in our accompanying video, are more complex and challenging than most presented in prior work, further highlighting the merit of our method.}

%% file: 02_related_work.tex
\section{Related Work}
\label{lr}
In this section, we discuss prior work relevant to traversability estimation for unstructured, vegetated environments. 
Traversability estimation or terrain assessment is the assessment of a patch of terrain in order to determine a ground robot's ability to enter, reside in, and exit that patch or volume without entering a failure state~\cite{papadakis2013terrain}. In the binary case this can be defined either as safe to traverse or not. In the continuous case, \ac{TE} can be seen as the likelihood of entering a failure state.
In the autonomous navigation literature, environments are characterized by the degree of structure humans impose on an environment. Structured environments are commonly man-made (e.g. offices, roads). Semi-structured environments are areas where some degree of structure is imposed, but there are uncontrolled elements, e.g. agriculture or driving on gravel roads (off-road). This work focuses on unstructured environments, where no human-imposed order or structure is introduced. Hence, research conducted in the area of autonomous navigation for agriculture, construction, or off-road driving is often not suitable for our case since these areas make strong assumptions about the environment based on the imposed structure. We refer the interested reader to broader surveys in these areas~\cite{borges2022survey, guastella2021learning}. 

To classify an environment as `vegetated' is inherently subjective.  In different works it can encompass natural areas devoid of overhanging foliage, a large area with a few trees~\cite{meng2023terrainnet} or an area with a cluster of trees in a park or forest void of undergrowth~\cite{frey2023fast}. 
Consequently, there is a large variation in works presented in the context of vegetated environments, or more specifically forests. Our video depicts an example of a scene from our data set that, compared to the literature, contains significantly denser vegetation with significantly more obstacles and clutter both at and above ground height.

\subsection{Geometric traversability estimation in vegetated environments}
Geometric \ac{TE} methods rely on reconstructing and maintaining a geometric representation of the environment using range sensors, RGBD cameras, or stereo-vision, usually assuming a rigid environment. The recent DARPA SubT Challenge required robots to navigate large-scale, complex underground environments and find and report object locations autonomously. The top two teams relied on geometric methods generating 2.5D cost maps~\cite{chung2023into}. The second-placed team generated 2.5D cost maps from a 3D probabilistic occupancy map and relied on classical heuristics, e.g. max slope angle and step height~\cite{HudTal21}. The winning team deployed a probabilistic GPU-based 2.5D elevation map with high grid resolution (\qty{4}{\cm}), and relied on learning-based \ac{TE}~\cite{miki2022elevation}. 2.5D representations, however, suffer in the presence of clutter and overhanging elements, as they cannot naturally represent them, which results in viable paths being closed off. 

\ac{TE} approaches in vegetated environments relying solely on lidar aim to capture the environment in 3D with additional salient features. Point-cloud-based approaches~\cite{lalonde2006natural, ahtiainen2017normal} have aimed to generate salient features, such as eigenvalue-features and slope inclination, without a probabilistic representation and have had little success. They suffer from high computational costs, low fidelity for discrimination, and noise from outliers. Probabilistic 3D voxel grids have been able to address the memory and computational issues at the cost of discretization~\cite{hornung2013octomap, saarinen2013normal, stepanas2022ohm}. Octomap~\cite{hornung2013octomap} efficiently models occupancy of the environment but lacks any additional salient features for accurate \ac{TE} in complex unstructured environments. 
Sarrin et al.~\cite{saarinen2013normal} introduced sub-voxel resolution occupancy mapping, which was subsequently expanded to a Normal Distributed Transform Traversability Map (NDT-TM)~\cite{ahtiainen2017normal}. NDT-TM introduced salient ray features, intensity, permeability, roughness, and slope angle to estimate \ac{TE} in vegetated environments at \qty{0.4}{\m} voxel resolution. Our prior work further enhanced this by fusing lidar multiple returns (or echoes), color, and adjacency features into the map representation~\cite{ruetz2022FTM}. This allowed for \ac{TE} in challenging vegetated environments for post-processed maps~\cite{ruetz2022FTM} at higher resolutions (\qty{0.1}{\m}). However, this method suffers from computational constraints and accuracy degradation during deployment, when sensor readings are received continuously on the robot rather than post-processed. Maintaining additional ray-based features per voxel comes at a high computational and memory cost, and is therefore rare for real-time systems at high resolution. 
Recent work has shown that \acp{SCNN} are suitable for \ac{TE} in complex structured environments by leveraging contextual information of the environment~\cite{frey2022locomotion} whilst being computationally efficient. The method was trained purely in simulation using 3D occupancy and requiring a large data set, equivalent to 57 years of real-world experience, a quantity that would be difficult and costly to obtain in real vegetated environments.

\subsection{Vision-based traversability estimation}
In scenarios with vegetation, geometric-only methods are commonly considered insufficient as they lack discrepancy. Hence, many studies have aimed to solve \ac{TE} in these environments using solely image modalities. These methods aim to assess \ac{TE} and plan from a single image (single viewpoint)~\cite{kahn2020badgr, wellhausen2019where, siva2022nauts}. Classically, they aim to classify the terrain into different semantic classes and assign each semantic class a traversability score. The concept of class drift and often fluid boundary between classes, e.g. bush vs bramble, can become challenging. However, they have shown high accuracy in complex situations but can be prone to reliability issues, e.g. sensor reading degradation or camera occlusions~\cite{bradley2015scene}, as they gather information from a single input and retain no memory.  

\subsection{Hybrid methods}
Late-fusion methods combine multiple representations that can stem from different sensor modalities or estimation processes, e.g. fusing semantic segmentation with a 2.5D elevation map. Maturana et al.~\cite{maturana2018real} demonstrated the fusion of semantics and 2.5D height maps for off-road (gravel road) navigation, avoiding vegetation where possible. Semantic and spatial cost maps (e.g. semantics, height difference, ceiling height) for high-speed off-road driving using stereo-vision has been successfully implemented~\cite{meng2023terrainnet}. The authors emphasize fast inference time at the cost of (geometric) representation fidelity and aim to avoid complex areas. Bradley et al.~\cite{bradley2015scene} demonstrated a \ac{TE} approach of separating the environment support surface and an above-ground point cloud classification using classical machine learning and semantic segmentation on a data set collected in different challenging environments. Similarly, an approach relying on semantic segmentation for traversability assessment and support surface estimation from an RGBD sensor has shown good results~\cite{li2023seeing}. Frey et al.~\cite{frey2023fast} demonstrated an incremental, online learning approach that combines a traversability signal with anomaly detection. Real-time operation is shown in natural environments, mostly parks and hiking trails with light vegetation. The authors claim a vision-only approach but project their image-based classification onto a 2.5D height map~\cite{miki2022elevation} to generate signed distance fields suitable for navigation. In many of these cases, the fusion of these methods relies on the underlying geometric representations as 2.5D height maps. Conversely, they are unsuitable for environments with significant vegetation and overhanging clutter, as shown in Fig.~\ref{fig:mcc_vs_vegd}. 

\subsection{Learning on 3D representations}
\review{Learning-based methods on 3D data is a growing field of interest. However, research lags behind the vision-based community's performance for general learning tasks such as classification and semantic segmentation. In general, methods are often developed in the 2D image domain and then adapted to 3D~\cite{camuffo2022recent}. The same recent review noted that challenges of learning on 3D representations arise from sparsity of data, lack of salient features compared to image modalities, large training data sets required, and large computational requirements. \ac{SCNN} allow for a reduction of computational time by ignoring empty voxels~\cite{liu2015sparse}. \ac{SCNN} using a U-Net architecture with skip connections has been shown to allow for the learning and embedding of features and surrounding context for different desired tasks, such as semantic instance segmentation~\cite{choy20194d},  multi-object classification~\cite{huang2020indoor}, scene completion~\cite{hoeller2022neural} and detection~\cite{gwak2020generative}. Compared to our proposed method these approaches rely on a combination of occupancy representation and/or (point- or voxel-wise) semantic labels, and not a rich ray-based probabilistic map. Further, we can leverage a small data set, data collected in a few hours, compared to the (simulated) experience equivalent to 57 real-time years in~\cite{hoeller2022neural, frey2022locomotion}.}

\subsection{Learning from Experience and data set}
\label{sec:realted_workd_lfe}
Generating accurately labeled data is essential for learning-based methods. For \ac{TE}, currently available simulators do not offer enough physical or perceptual accuracy to replace real-world data gathering. Gathering self-supervised data by associating the robot's state with sensory data to produce high-quality ground truth data is referred to as \ac{lfe}~\cite{wellhausen2019where, ahtiainen2017normal, kahn2020badgr, frey2022locomotion}. The positive, traversable case, where the robot successfully traversed, can be easily obtained and is commonly used. However, gathering non-traversable examples can be hazardous and costly, potentially damaging the robot, and hence, heuristics or hand-labels are typically favored in the literature. 
In contrast to the works above, in this paper we present a principled approach that probabilistically combines hand labels with robot experience to generate high-quality labeled data. 

Additionally, few data sets are available in the target domain~\cite{jiang2021rellis, wigness2019rugd, bae2023self}. RUGD~\cite{wigness2019rugd} contains a combination of images of natural environments, whereas Rellis3D~\cite{jiang2021rellis} generates labeled point cloud data with image-to-lidar projection, annotating natural elements as ``grass'', ``bush'' and ``tree''. A pure point cloud data set was announced but, to the best of the authors' knowledge, not yet released by Bae et al.~\cite{bae2023self} containing point-wise labels in off-road environments void of dense underbrush or challenging vegetation. A data set containing ray-based features and labels is not available.

Our paper builds on existing literature's findings by leveraging a rich probabilistic map representation and the \ac{SCNN}'s ability to incorporate environmental context to form a novel, high-performing, real-time capable system. Our work further differentiates itself from the SOTA by only using a limited amount of real-world data in combination with robot experience. The resulting method is capable of performing accurate \ac{TE} in environments more challenging than previously demonstrated by other methods.

%% file: 03_methodology.tex
\section{Methodology}

\begin{figure*}[t]
    \centering
    \includegraphics[width=\textwidth]{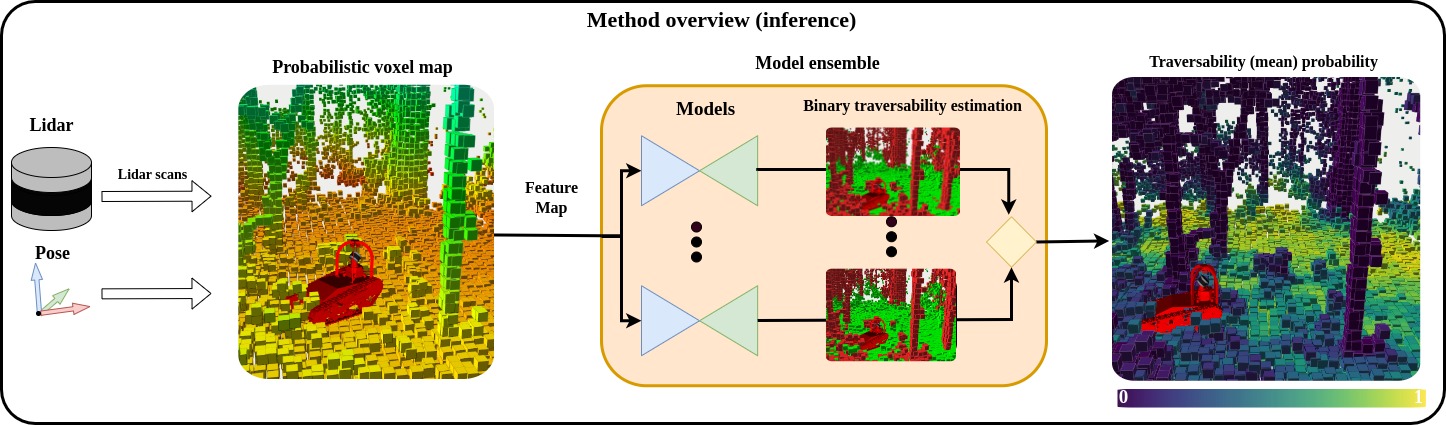}
    \caption{\review{Method overview (inference): During robot deployment a continuous stream of lidar measurements are continuously fused into a single probabilistic 3D voxel map, representing the environment with per-voxel statistics. A local feature map is generated to assess the traversability of a local area around the robot's current pose. Each of the $N$ models independently classifies voxels as either traversable or non-traversable. The ensemble creates the traversability probability for each voxel by taking the mean of the $N$ binary classifications. A sample scene is shown on the top left, with the robot in red. }}
    \label{fig:methodoverview}
\end{figure*}

\subsection{Problem Definitions and method overview}
\label{method:problem_description}
\review{ The goal of this work is to generate a full 3D \ac{TE} map suitable for allowing an \ac{AGV} to navigate in natural, densely-vegetated environments. This work represents the environment with a probabilistic voxel map $\textbf{M}$, containing non-overlapping 3D box volumes where each voxel $m$ retains ray statistics. Further, we assume each voxel \mbox{$m \in \textbf{M}$} has a true binary traversability state $\tau \in \left\{TR, NT\right\}$ where $TR=1$ and $NT=0$ represent traversable and non-traversable states respectively. This work uses a supervised learning approach to learn a parametric function $ f_\Theta(x) $ that can estimate $\tau$ for each voxel from the input sample \textbf{x} described below, formally  $\tau  = f_\Theta(x) $.  This work aims to learn this parametric model that can accurately assess $\tau$ on a map $\textbf{M}$ that is continuously updated whilst the robot is moving and sensing the environment.}

\review{ Fig.~\ref{fig:methodoverview} provides a high-level overview of the proposed method for the inference step, allowing for online \ac{TE}. A continuous stream of lidar measurements is fused into a 3D probabilistic voxel map in a static global frame during robot operation. The map is initialized with no prior values at the start of a deployment and is progressively updated as the robot traverses the environment and records sensor data. The map internally tracks multiple lidar ray statistics for each voxel through raycasting; commonly at \qty{0.1}{\m} resolution. Given sufficient measurements, the statistics within the voxel itself are assumed to converge and adequately represent the underlying environment, further discussed in Section~\ref{subsec:prob_map}.  During deployment, a local region is extracted from the global probabilistic map at fixed time intervals. The local feature map contains all voxels and their statistics for a box volume centered around the current robot pose. The voxel statistics are used directly as the input features to the prediction network, avoiding any additional feature calculation, see~\ref{subsec:prob_map}. The local feature map is then passed to an ensemble of prediction networks. The ensemble consists of $N$ binary traversability classifiers (U-Net models) trained on the same data from different (random) weight initializations. Each model estimates a binary traversability state for each output voxel, generating a binary traversability map at the same size and resolution as the input map. In Fig.~\ref{fig:methodoverview} red voxels are non-traversable and green voxels are traversable. Finally, a per-voxel traversability probability is calculated by averaging the binary traversability estimate of the $N$ models. We use this deep ensemble to increase robustness against outliers and generalize performance to novel environments. Ensembles have proved to be a popular choice for generating probabilistic estimates and uncertainty quantification from neural networks~\cite{lakshminarayanan2017simple}, although we do not provide calibration scores.}

\begin{figure*}[t]
    \centering
    \includegraphics[width=\textwidth]{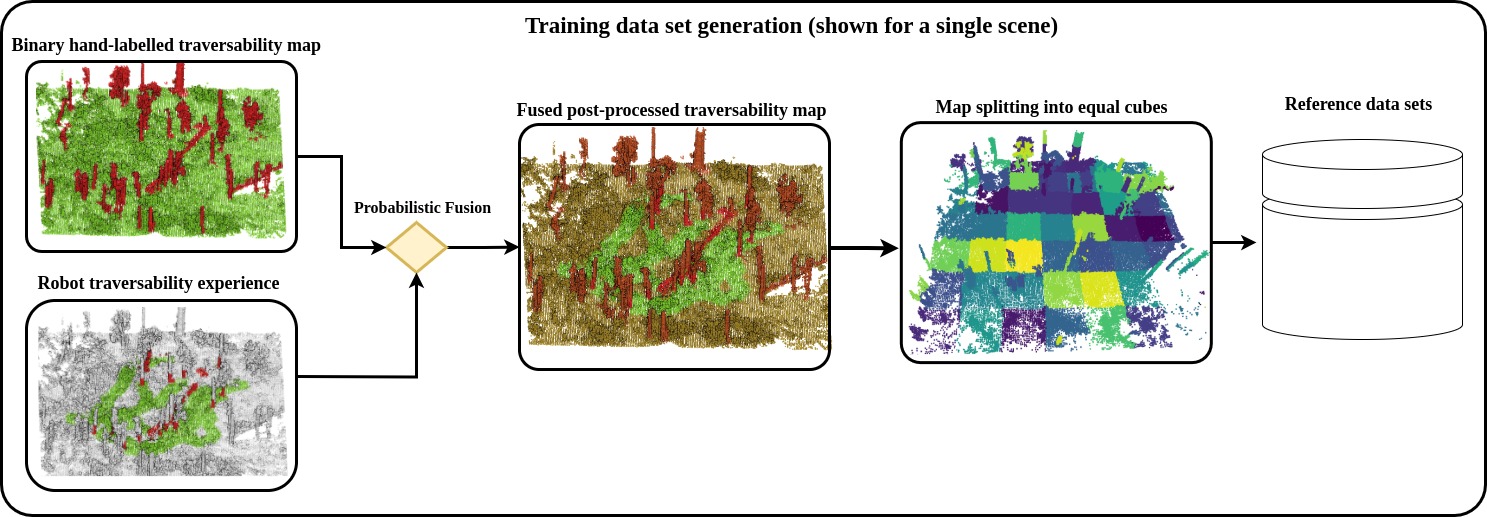}
    \caption{\review{Training data generation: The training data is generated based on the post-processed map, using offline optimized poses. The hand-labeled data set is fused with the robot experience. The fused post-processed traversability map is split into smaller cubes suitable for training our method. This is repeated for each scene and added to the reference database, containing all data for training. The exception is the separate test set, where the scene is not split into smaller cubes. Details of each step are provided in Secs. C-F}}
    \label{fig:data_set_generation}
\end{figure*}

Fig.~\ref{fig:data_set_generation} illustrates the offline training by splitting $K$ post-processed voxel maps into the training data set. Each map is split into equally sized 3D cubes. This process is illustrated for a single map, and each cube is visualized with a different color. All the cubes of all the $K$ maps are combined into a single training data set. The test data is a separate map evaluated without splitting it. A detailed description is provided for the used data set (Sec.~\ref{method:data_set}), and the training details (Sec.~\ref{method:trainning}).

\subsection{Probabilistic 3D Environmental Representation}
\label{subsec:prob_map}
The mapping process aims to estimate each voxel's state based on a continuous stream of lidar observations. The state is a set of different ray-based probability distributions assumed to represent the environment given sufficient sensor readings. Our approach uses \ac{NDT-OM}~\cite{saarinen2013normal} to capture sub-voxel resolution geometric distributions and extends the number of statistics to allow for accurate traversability estimation in dense forest environments. More specifically, each voxel $m$ contains: a 3D multi-variate Gaussian distribution with probabilistic occupancy information characterized by the mean position $\mu_{NDT}$ and covariance $\Sigma_{NDT} = S_{NDT} * S_{NDT}^T$ (stored as the triangular square root covariance matrix $S_{NDT} \in \mathbb{R}^6$ for computational efficiency); the number of rays that ended within this voxel $N_{OCC}$; the occupancy probability in log-odds form $l_{OCC}$; and the ``hit'' and ``miss'' counts of the rays that have statistically ended or passed through the \ac{NDT} distribution with a close enough margin. These voxel statistics have been previously used for permeability calculation in~\cite{ahtiainen2017normal}. In addition, we also store in each voxel: the lidar intensity mean $\mu_{INT}$ and variance $\sigma_{INT}$, which is relevant to chlorophyll-rich elements~\cite{ahtiainen2017normal}, and the number of multi-returns $N_{MR}$ which occur when a lidar beam is partially split from thin elements or edges, for example. Note that an increase in the number of second returns can be observed in voxels containing multiple small vegetation elements such as grass blades, leaves and thin stems.

We rely solely on the values characterizing the distribution for the proposed method. This avoids computationally expensive additional feature calculations and allows us to learn the feature embedding directly. An input feature $\bm{x} \in \mathbb{R}^{13}$ is defined as 
\begin{equation}
\begin{split}
    \bm{x} = [ S_{NDT}, N_{OCC}, l_{OCC}, \mu_{INT}, ... \\ \sigma_{INT}, N_{HIT}, N_{MISS}, N_{MR} ]
\end{split}
\end{equation}

\review{Our previous work explored the relevancy of features calculated from these statistics~\cite{ruetz2022FTM}}. We have developed our online mapping method on top of \ac{OHM}~\cite{stepanas2022ohm}, an open-source 3D probabilistic mapping framework developed with performance in mind. This allows us to maintain numerous salient voxel statistics during online deployment. No other 3D probabilistic mapping framework is currently available to provide such feature-rich and real-time capable 3D representations. 

\subsection{Data set generation}
\label{method:data_set_generation}

\review{Our method relies on two different processes to generate the probabilistic map representation. All the maps are generated from recordings of real-world data, recorded as the raw sensor readings of a robotic platform, see Sec.~\ref{sec:res_platform}}. 

\review{The offline, post-processed probabilistic map \mbox{$M_{Post}$} is generated from all data from a particular experiment. Using an optimized global, post-processed trajectory to minimize drift with the SOTA Wildcat inertial-lidar SLAM algorithm~\cite{ramezani2022wildcat}. This post-processing pipeline can handle long deployments and fuse data on large-scale maps on a consumer-grade laptop without needing to release or remove any data; the map is unbounded. The offline, post-processed map generates training and test data sets. }

\review{
The online-generated probabilistic map $M(t)$ for a particular experiment is the state of the probabilistic map at time $t$, given all previous sensor measurements and pose estimates from the start of the experiment $t_0$ up to $t$. This map is used for the online inference and qualitative evaluation presented in Sec.~\ref{sec:qualative_eval} and the accompanying video. To ensure online real-time capability, a fixed-size local map around the robot is maintained due to memory and computation constraints. Compared to the post-processed map \mbox{$M_{Post}$}, \mbox{$M(t)$} uses a local, online SLAM estimate for the pose estimate and may suffer from drift. Hence, to compare \mbox{$M_{Post}$} and \mbox{$M(t)$}, \mbox{$M(t)$} is aligned to \mbox{$M_{Post}$} using the iterative-closest point algorithm before calculating error statistics~\cite{chen1992object}. Some alignment errors may persist between local \mbox{$M(t)$} maps and \mbox{$M_{Post}$}, leading to the incomplete coverage statistics shown in Fig.~\ref{fig:results_temporal_performance}.
}

During robot deployment, \acp{AGV} commonly need to move into previously unobserved spaces. Initially, due to the limited observations, little information about that new area is available, and the quality of the map is low. The map quality temporally evolves and improves with the availability of additional sensor measurements and often by the \ac{AGV} moving. Therefore, the performance of the \ac{TE} depends on the quality of this spatially and temporally evolving map. Additionally, learning-based methods are commonly trained on post-processed maps, which have the highest possible quality. Hence, it is important to understand the performance implications of a method with regard to these temporal effects. This may arise due to distribution shifts between training and data available during deployment. This can lead to reduced accuracy, as was observed in~\cite{ruetz2022FTM}. In this work, we additionally aim to understand how resilient an estimator's performance is to map quality. 

The term ``map quality'' is a conceptual term that reflects how well the statistical distribution in each voxel represents the true environment and how well it has been observed. Voxels and associated feature statistics are assumed to be spatially independent, and combining different features or statistics is a common practice for data-driven and learning-based approaches. We treat time as a proxy for map quality and assume that the map improves as more data is gathered.

\subsection{Probabilistic collision maps for learning from experience}
\label{method:collision_map}
\begin{figure}
    \centering
    \includegraphics[width=\columnwidth]{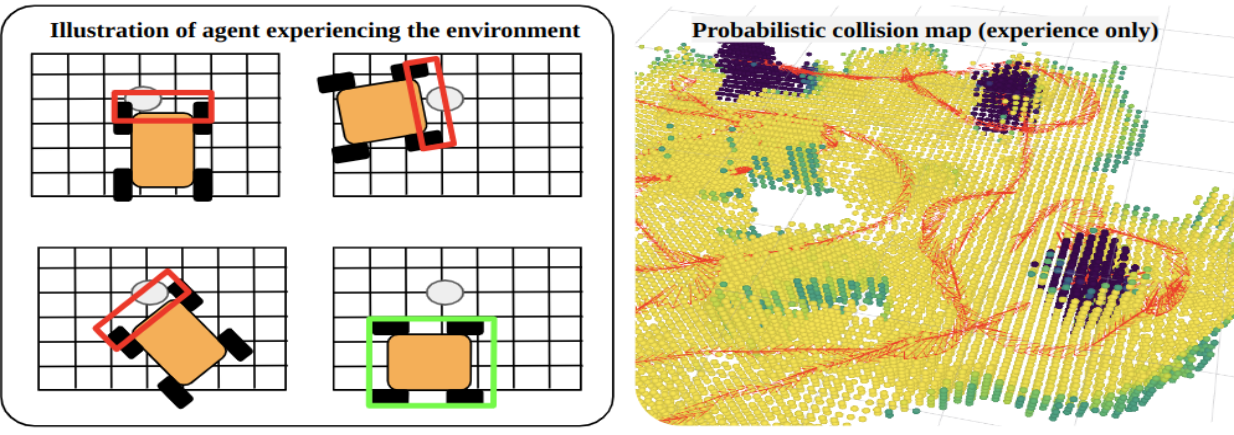}
    \caption{Left: Illustration of instances of a robot traversing or colliding with environmental elements. The red bounds indicate the voxels that may cause collisions, green boxes are voxels that the robot successfully traversed. Right: The probabilistic collision map from the robot experience only. Dark purple is non-traversable, yellow traversable and green uncertain. The red arrows are discrete poses of the trajectory. The map correctly captures the voxels mostly likely to be responsible for the collisions (tree trunk) and the adjacent traversable grass, without any discretization effects.}
    \label{fig:collision_map}
\end{figure}

Any supervised data-driven approach for \ac{TE} requires accurately-labeled data. As noted in previous work, labeling the traversability of densely vegetated environments can be particularly challenging in practice. Therefore, in this work we combine learning from robot experience (LfE) and human domain knowledge to generate accurately labeled data in a post-processing step. An additional collision layer is added to the map $\textbf{M}_{Post}$ that already contains the exteroceptive data. The collision layer tracks for each voxel $m$ the likelihood of being part of a collision $l_c(m |c_{1:t}, T_{1:t})$ and thus non-traversable. The trajectory $T_{1:t}$ and collision states $c_{1:t}$ are the sequence of recorded robot poses $T_t$ and collision states $c_t \in \{TR, NT\}$ at time $t$. A stationary binary-Bayesian filter in its log odds formulation is used to process the robot experience and update the voxels recursively. Compared to Octomap~\cite{hornung2013octomap}, we use the robot's pose and its collision state as the observation model instead of a range sensor, recording the robot's experience. For a voxel $m$ the update is performed as:
\begin{equation}
    l(m | c_{1:t}, T_{1:t}) = l(m_i | c_t,  T_{t}) +l(m | c_{1:t-1}, T_{1:t-1})
\end{equation}
If the robot is in a collision-free state $c = TR$, then all voxels $m$ within the robot's bounding box are updated. In the case of a collision $c = NT$, the voxels within the bounding box at the front of the robot and a threshold distance beyond (\qty{0.2}{\m}) are observed as non-traversable. The rationale is that the obstacles often extend beyond the voxel containing the object the agent directly interacts with, e.g. tree trunk or thick bramble bush. The distance threshold value was found heuristically and depends on the environment, not the voxel resolution itself. This observation model assumes that collisions are due to the elements in front of the agent and prevent it from advancing when the robot moves predominantly forward. During the collection of the data set for this work, this was enforced by the operator turning only in safe areas and primarily moving forward during data collection. A visualization is shown in Fig.~\ref{fig:collision_map}.
In practice, hand-labeled data is used to initialize the collision layer with fixed prior probabilities for each traversability class with empirically selected values of $p_{NT} = 0.3$ and $p_{TR} = 0.7$. During data collection, an expert remotely operates the robot through vegetated environments and manually records collision events on the handset. The experience of the robot overrides hand-labeled data. The rationale is that the experience of the robot itself, failing or succeeding to traverse a region, is what we are trying to label. Issues arising from discretization are addressed using the probabilistic updates. Hence, this method allows us to generate high-quality data for our learning approach combining expert domain knowledge and refining it with robot experience in a principled fashion.

\subsection{Network Architecture and Training Procedure}
\label{method:trainning}
\begin{figure}[t]
    \centering
    \includegraphics[width=\columnwidth]{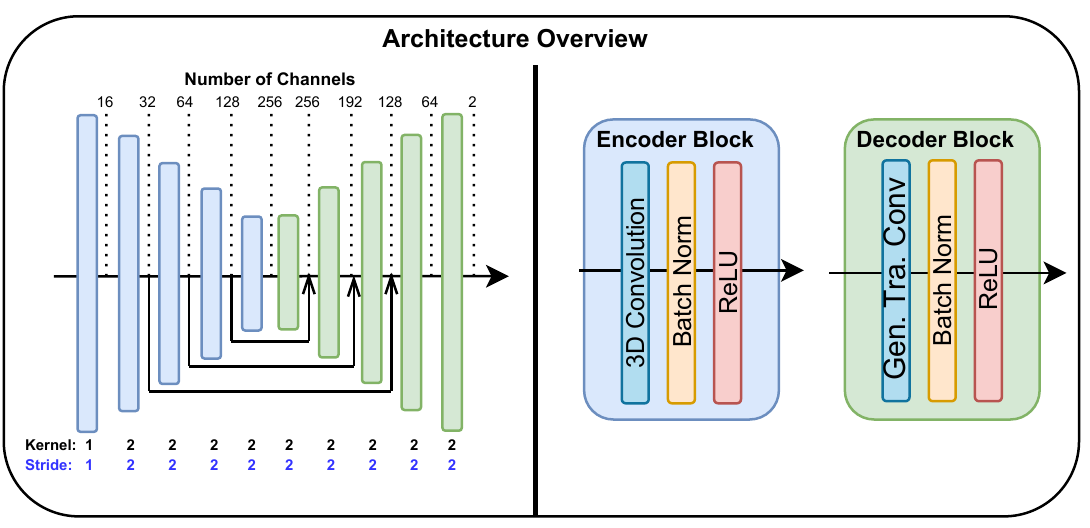}
    \caption{Overview of the U-Net architecture used in this work showing the number of channels, kernel sizes, strides, and skip connections.}
    \label{fig:net_architecture}
\end{figure}
We employ a 5-layer U-Net architecture~\cite{ronneberger2015u} (see Fig.~\ref{fig:net_architecture}). \review{ This architecture has been shown to be suitable for the tasks of classification, scene completion and TE in structured environments on sparse voxelized 3D data for robotic applications~\cite{hoeller2022neural}. Comparatively, we differ from this work by leveraging a much richer feature set per voxel, where~\cite{hoeller2022neural} relies on either occupancy and semantics or a combination of both.} The resulting network is relatively small ($\sim$2M parameters versus standard U-Net $\sim$30M). We use cross-entropy loss from the TorchSparse implementation of \acp{SCNN}~\cite{tang2022torchsparse}.

To combine the different post-processed scenes or maps into a data set format suitable for training, we split the maps into smaller cubes of $32\times 32 \times 32$ voxels. Each voxel appears in only one cube, and a cube requires at least $n=150$ non-empty voxels to be considered valid. These map cubes are then concatenated into a single large data set. We employ a classical ten-fold cross-validation with a held-out test scene (\#9) ensuring a fair comparison against other methods. The input features are individually scaled using zero-mean normalization, with the scaling calculated based on the training set, excluding the validation data set. Different data augmentation methods are applied without violating the voxelization. To introduce noise, a \qty{5}{\percent} pruning chance is applied on a voxel-based level. Geometric augmentations are applied on a cube level, including mirroring, translation, and fixed angle rotation ($\phi \in \left\{0, 90, 180, 270\right\}^\circ$ around the gravity-aligned axis with a uniform chance). The fixed rotation intervals are required to maintain alignment with the voxel representation. In addition, a random translation of 1-10 voxels is applied to each cube with a \qty{50}{\percent} chance in any axis-aligned direction.

\subsection{Data Set}
\label{method:data_set}
\begin{figure*}
    \centering
    \includegraphics[width=\textwidth]{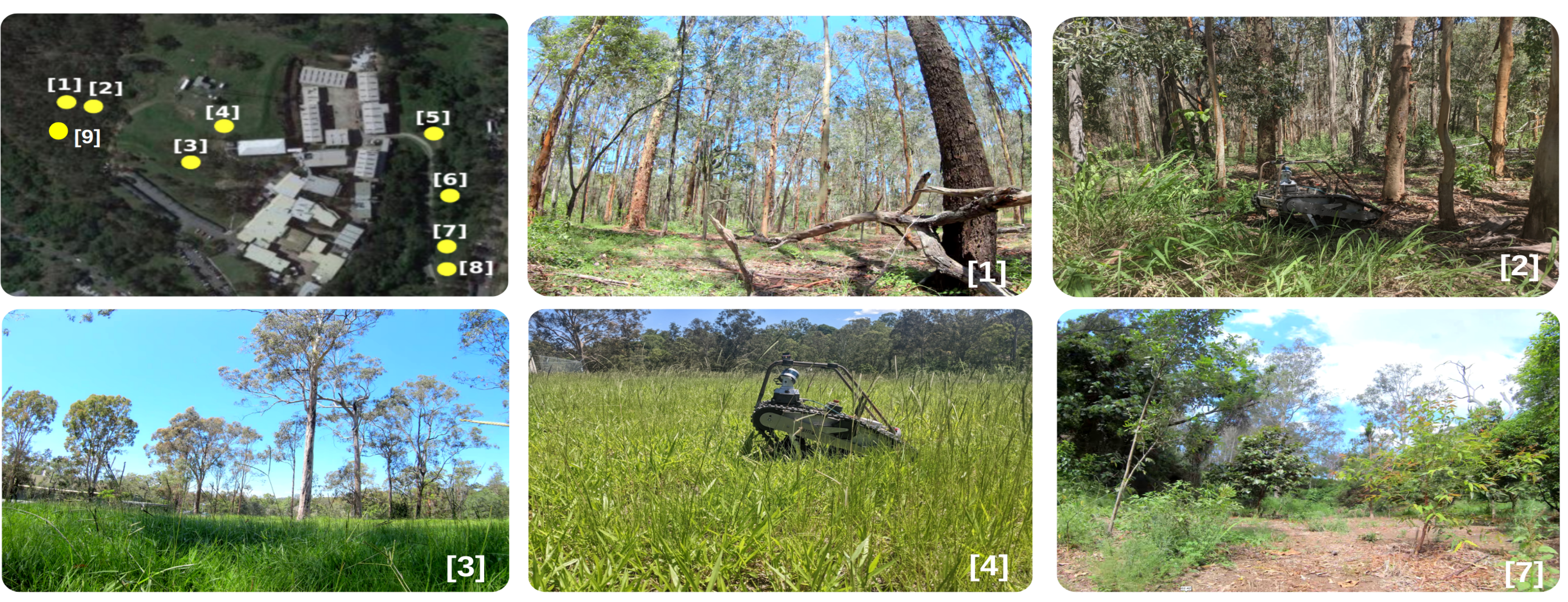}
    \caption
    {Top far left: QCAT testing facility in Queensland, Australia and the approximate locations where the data sets were gathered. [1] and [2]: robot in forest environments from the robot and external viewpoint, respectively. [3] and [4]: robot in tall grass with some large trees. [7]: environment with a mixture of small to medium trees and patchy vegetation.}
    \label{fig:data_set_environments}
\end{figure*}

The data set generated as part of this work has been open-sourced\footnote{Data sets available at \url{https://data.csiro.au/collection/csiro:58941} }, and Table~\ref{tab:data_set_overview} provides the statistical summary. 
The data set contains a total of 1,016,243 points and covers approximately \qty{3260}{\m^2} from nine different vegetated environments. Each point contains raw distribution values of the layers, the \ac{TE} labels, and the ray features from \ac{NDT-TM} and \ac{FTM} ~\cite{ahtiainen2017normal, ruetz2022FTM}. This allows this data set to be useful for future novel methods and easy comparison. In addition, this data set contains high-quality hand labels directly generated on the point cloud itself, avoiding miss-labeling elements due to occlusion and clutter in the vegetated environment when using point-to-image projection.
The data was collected by an expert initially tele-operating an \ac{AGV} in vegetated environments and recording collision states. Using the described method above~\ref{method:collision_map}, the exteroceptive data was post-processed into \mbox{$M_{Post}$} and point clouds. The resulting point cloud was hand-labeled for traversability class by an expert with domain knowledge. The different data sources,  \mbox{$M_{Post}$}, hand-labeled data, and collision maps were finally fused, as described in Sec.~\ref{method:collision_map}. This produces a high-quality data set, and visualizations are shown in Fig.~\ref{fig:data_set_environments}. 

Table~\ref{tab:data_set_overview} summarizes the environments and data sets by scene at \qty{0.1}{\m} voxel resolution. The location of the scenes can be seen in the bottom right image of Fig.~\ref{fig:collision_map}. The second column shows the total number of labeled voxels $N_{labels}$. Columns three to six show the percentage of labels that are traversable ($TR$) or non-traversable ($NT$), and the labeling method; hand-labels (HL) or learning from experience (LfE). 

The seventh column shows the mean \ac{vd} per scene. In this work, we define \ac{vd} as a column-based metric that aims to capture how much of the volume above the ground, and including a patch of ground, contains measurements relevant for TE for the robot. We use \ac{vd} as a proxy for vegetation density as our environment only contains vegetation. Further, we only consider elements up to a meter above the ground. This is the intersection most \ac{AGV}s will interact with and for which they need to reason about the vegetation pliability. Elements above that height, such as the dense foliage of the tree, can skew the density metric. For each column of voxels, the set of voxels starting from and including the ground point up to a threshold of \qty{1}{\m} are considered. For that column, \ac{vd} is the ratio of the number of voxels containing measurements $(N_{meas})$ over the total number of voxels in the set $(N_{total})$; $\rho_{V} = \frac{N_{meas}}{N_{total}}$. It is free from sensor models and can be calculated on voxelized or raw point clouds and can provide a comparison metric against other data sets. A higher number indicates greater density in a scene hence more potential traversability obstacles. 
The highest value of \ac{vd} corresponds to the presence of vegetation in every voxel of the column from the ground to \qty{1}{\m} height, while the lowest value corresponds to the presence of vegetation only at the ground (up to a height nearing the size of the voxel). 
Lastly, we provide the dimensions of the bounding volume for each scene in $x,y,z$ gravity-aligned coordinate frames, the default reference frame for our SLAM system. 
     
 \begin{table}[]
    \caption{Overview of the data set}
    \resizebox{\columnwidth}{!}{
    \begin{tabular}{llllllll}
    \toprule
Scene & $N_{voxels}$ & $TR$ HL [\%] & $TR$ LfE  [\%]& $NT$ HL  [\%]& $NT$ LfE  [\%] &  Density & Dimensions [m] \\ \hline
\# 1 &   86613 &  0.45 &  0.22 &  0.29 & 0.04 & 0.52 & $ 18.2 \times 12.0 \times 3.9 $ \\
\# 2 &   88410 &  0.6  &  0.2  &  0.16 & 0.03 & 0.41 & $ 17.5 \times 16.0 \times 3.5 $ \\
\# 3 &  146121 &  0.66 &  0.32 &  0.02 & 0    & 0.52 & $ 24.6 \times 15.7 \times 4.3 $ \\
\# 4 &  282197 &  0.62 &  0.38 &  0    & 0    & 0.55 & $ 30.0 \times 26.0 \times 3.4  $\\
\# 5 &  113288 &  0.53 &  0.47 &  0    & 0    & 0.46 & $ 23.3 \times 17.9 \times 2.5 $ \\
\# 6 &   73073 &  0.26 &  0.18 &  0.44 & 0.12 & 0.57 & $ 13.3 \times 15.0    \times 1.9 $ \\
\# 7 &  120442 &  0.47 &  0.22 &  0.26 & 0.04 & 0.46  & $ 25.1 \times 18.8 \times 1.8 $ \\
\# 8 &  102560 &  0.48 &  0.09 &  0.4  & 0.02 & 0.48 & $ 14.8 \times 17.7 \times 3.0 $ \\
\# 9 &   31539 &  0.67  &  0.25  &  0.06 & 0.02 & 0.4 & $ 10.6 \times 10.3 \times 2.6 $  \\
    \bottomrule
    \end{tabular}
    }
\label{tab:data_set_overview}
\end{table}

%% file: 04_experiments_results.tex
\section{Experiments and Results}
\label{sec:results}
In this section, we provide a comprehensive experimental evaluation of our proposed method.
The first subsection gives an overview of the platform and implementation used in this study. The second details our method's experiments, contrasting them with the SOTA. The third offers an ablation study to understand performance, while the fourth analyzes the estimator's robustness to map quality. The final subsection illustrates qualitative examples using our open-source forest data set. 

\review{We evaluate the performance of the proposed method using the \acf{MCC}~\cite{chicco2020advantages}. \ac{MCC} is a metric for assessing binary classification that is suitable for imbalanced data sets. Unlike the F1 score, it considers all four classification cases (TP, FP, TN, FN), making it robust against class imbalance and offering invariance to class swap~\cite{chicco2020advantages}. Valid \ac{MCC} scores range from -1 to 1, where 0 is a random chance, 1 is perfectly correlated, and -1 is a negatively correlated model. Other commonly used classification metrics such as the F1-score, precision, and recall, overestimate an estimator's performance in cases of high class imbalance and depend on the choice of the positive class. In terrain traversability estimation, the common convention is to choose the traversable class as the positive class, which is typically also the majority class. Hence, we argue that using the F1 score alone may not be suitable for this application. Nonetheless, the F1-score is provided in this paper along with the MCC to allow for some comparison with the SOTA. In the cases where cross-validation is used, the mean $\mu$ and standard deviation $\sigma$ are reported to provide insight into how well the models are trained and generalized.}

\subsection{Experimental Platform and Implementation Details}
\label{sec:res_platform}
Data was collected with a Dynamic Tracked Robot (DTR), a \qty{35}{\kg} tracked vehicle equipped with a sensory pack containing IMU, a rotating Velodyne VLP-16 lidar angled at \qty{45}{\degree} and four RGB cameras. The lidar is mounted on the robot and performs a full revolution at \qty{2}{\Hz}. Fig.~\ref{fig:intro_img} depicts the vehicle in an environment representative of this work. Further details on the platform can be found in~\cite{ruetz2022FTM}.

The following learning parameters are used for training our method: learning rate $l_r = 10^{-4}$, weight decay $w = 9\times 10^{-4}$, batch size $b = 64$, early stopping patience $es = 5$ and maximum epoch number $epoch_{max} = 250$, with average convergence between $50-100$ epochs. \review{The ADAM optimizer was used to train the models. The loss calculated on the validation set was used as the early stopping criterion. Hyperparameters were tuned using a grid search and choosing the settings with the lowest MCC-score variation for a ten-fold cross-validation only of the training data. To compare the different methods, a ten-fold-cross validation was used with a hold-out test set. Model weights are generated, stored and retrieved using the Python PyTorch library ~\cite{paszke2017automatic} with the TorchSparse extension~\cite{tang2022torchsparse}}. 

Training time was approximately \qty{210}{\second} per model onboard a Precision 7750 laptop equipped with a Quadro RTX 5000 Mobile Max-Q, \qty{64}{\giga\byte} of system memory, and a \qty{2.30}{\GHz} Intel Core i7-10875H CPU. 
On this hardware, the ensemble of 10 models processes each local map at \qty{0.1}{\m} voxel resolution at \qty{3}{\Hz}  for a \qtyproduct{10 x 10 x 2}{\m} scene and above \qty{1}{\Hz} for a \qtyproduct{20 x 20 x 5}{\m} scene. 

\subsection{Evaluation}
\subsubsection{Accuracy and comparison to SOTA 3D methods}
\label{res:comparsion_sota}
We evaluate our proposed method against SOTA TE methods that utilize 3D voxel representations and range sensors. Our comparison includes the classical Constant Threshold Classifier (CTC), which relies on the \ac{NDT} representation and leverages heuristics of roughness and slope angles~\cite{stoyanov2010path}. The CTC method was tuned on the test set to demonstrate the best possible performance heuristics achievable by this method in an ideal scenario; $\theta_{threshold} = \qty{30}{\degree}$ and $\rho = 0.001$.
Further, we compare against \ac{NDT-TM}~\cite{ahtiainen2017normal}, and \ac{FTM}~\cite{ruetz2022FTM}, two SOTA methods that leverage 3D probabilistic voxel representations that tackle TE in vegetated environments. \ac{NDT-TM} and \ac{FTM} were trained as described in~\cite{ruetz2022FTM}, and the reported results were reproduced to ensure a fair comparison. \review{All methods were tuned as best as possible and were evaluated and compared on the unseen test set, Scene \#9.} This aligns with common machine learning practices to ensure fair evaluation and comparison. In addition, it avoids possible domain knowledge leakage during training, a known problem with cross-validation approaches. Results for voxel sizes of \qty{0.1}{\m} and \qty{0.2}{\m} are presented. Lower resolutions are considered too coarse for the cluttered environment. For online deployment, higher resolutions are significantly more computationally expensive and provide a diminishing return in discriminating the environment, particularly since \ac{NDT} allows for sub-voxel resolution. Additionally, they require denser lidar scanning (more lidar beams) to ensure that enough laser rays traverse each voxel, ensuring the convergence of voxel statistics.

\begin{figure*}[t]
    \centering
    \includegraphics[width=\textwidth]{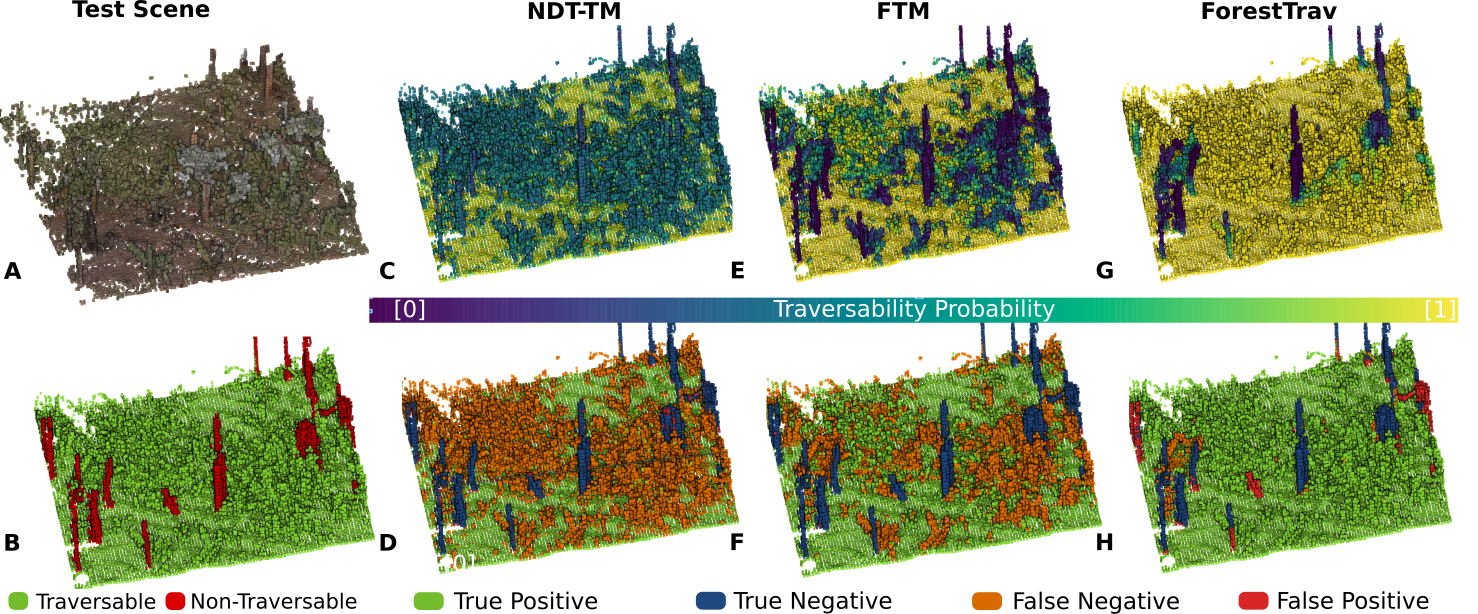}
    \caption
    {Left column: \textbf{A)} RGB voxel cloud and \textbf{B)} ground truth labels of the test scene. Second column: \textbf{C)} traversability probability of NDT-TM and \textbf{D)} classification results compared to the ground truth. Green is true positive (TP), blue is true negative (TN), orange is false negative (FN) and red is false positive (FP). The third and fourth columns show the same for FTM~\cite{ruetz2022FTM} (\textbf{E}, \textbf{F}) and ForestTrav(ours)  (\textbf{G},\textbf{H}).}
    \label{fig:comparsion_te_estimation}
\end{figure*}

\begin{table}[]
    \centering
    \caption{ForestTrav classification performance compared to the SOTA}
    \resizebox{\columnwidth}{!}{
        \begin{tabular}{lllcc}
        \toprule
 Method name & $vs$ [m] &   MCC $ (\mu \pm \sigma)$  & F1-Score $ (\mu \pm \sigma)$   \\ \hline
        CTC~\cite{stoyanov2010path}                               & 0.1 & $ 0.27 \pm  0.0 $ & $  0.46 \pm  0.0 $ \\
        NDT-TM~\cite{ahtiainen2017normal}               & 0.1 & $ 0.23 \pm  0.00$ & $  0.46 \pm  0.03 $ \\ 
        FTM~\cite{ruetz2022FTM}                         & 0.1 & $ 0.41 \pm  0.00 $ & $  0.61 \pm  0.04 $  \\
        \textbf{ForestTrav (Ours)}                      & \textbf{0.1} & $\bm{ 0.62 \pm  0.03 }$ & $\bm{ 0.82 \pm  0.04 }$  \\
        \hline
        CTC~\cite{stoyanov2010path}                              & 0.2 & $ 0.23 \pm 0.0 $ &$0.49 \pm 0.0 $ \\
        NDT-TM~\cite{ahtiainen2017normal}               & 0.2 & $ 0.30 \pm  0.00 $ & $  0.54 \pm  0.00 $ \\ 
        FTM~\cite{ruetz2022FTM}                         & 0.2 & $ 0.38 \pm  0.05 $ & $  0.59 \pm  0.00 $  \\
        \textbf{ForestTrav (Ours)} & \textbf{0.2}       & $\bm{ 0.49 \pm  0.04 }$ & $\bm{ 0.73 \pm  0.02 }$  \\
        \bottomrule
        \end{tabular}
    }
    \label{results:tab_method_comparison}
\end{table}
Table~\ref{results:tab_method_comparison} summarizes the results across methods. It shows that our proposed ForestTrav method significantly increases TE classification accuracy over other 3D voxel-based techniques, with \ac{MCC} increasing from 0.41 to 0.62 over the best performing alternative method (FTM) for \qty{0.1}{\m} voxel resolution and 0.38 to 0.49 for \qty{0.2}{\m} resolution. F1 scores obtained by ForestTrav are also significantly increased, reaching a high-performing 0.82 at \qty{0.1}{\m} resolution compared to 0.61 for FTM.
The relative comparative performance between the SOTA methods aligns with reported observations in~\cite{ahtiainen2017normal, ruetz2022FTM}. 
At higher resolutions, the geometric method CTC outperforms \ac{NDT-TM}. Note the initial results in~\cite{ahtiainen2017normal} were reported at much lower voxel resolutions (\qty{0.4}{\m}).

Fig.~\ref{fig:comparsion_te_estimation} A and E illustrate colorized, voxelized point clouds with the ground truth of the test set at a \qty{0.1}{\m} voxel resolution, encompassing open spaces and various vegetation densities. Significant vegetation near the ground, a typical challenge in forest navigation, must be considered for safe robot traversal. \ac{NDT-TM} accurately classifies all non-traversable elements but fails to distinguish pliable vegetation near the ground, mislabeling them as non-traversable (orange). In contrast, \ac{FTM} exhibits better discrimination of traversability probability but similarly misclassifies significant vegetated elements, albeit less often. However, neither method would allow an \ac{AGV} to traverse this environment.
ForestTrav generally identifies traversable and non-traversable elements accurately and would allow the robot to navigate safely in the scene. However, it misclassifies a central small tree trunk, the foot of some tree trunks and some trees on the scene's edges as traversable. We view the foot of the trees as a low concern, given that many voxels above are correctly classified as non-traversable, which would prevent planning through the tree. The trees on the edges of the scene are believed to be border effects occurring where the scenes were cropped. We have verified that these elements are correctly classified with more context (not part of the cropped test scene). Furthermore, we note that the model's traversability probability estimations are close to either extreme (0 or 1), with few values in the middle range. Ensembles are generally used as a technique to generate probabilities and epistemic uncertainty from neural networks, but can be overconfident in their predictions~\cite{opitz1999popular}.  

\subsubsection{Performance for different vegetation densities}
In this experiment we aim to understand if the three methods show differences in performance with respect to the amount of vegetation present, represented by the proxy \ac{vd} introduced in Sec.~\ref{method:data_set}. \ac{MCC} summary statistics to make these assessments were computed for test-scene \#9. All voxels were associated with a \ac{vd} score based on the column they fall into, even those above \qty{1}{\m}.  Points in voxel above \qty{1}{\m} are included in the evaluation to keep the same baseline of comparison as in results previously presented in Sec.~\ref{results:tab_method_comparison}.
\begin{figure}[t]
    \centering
    \includegraphics[width=\columnwidth]{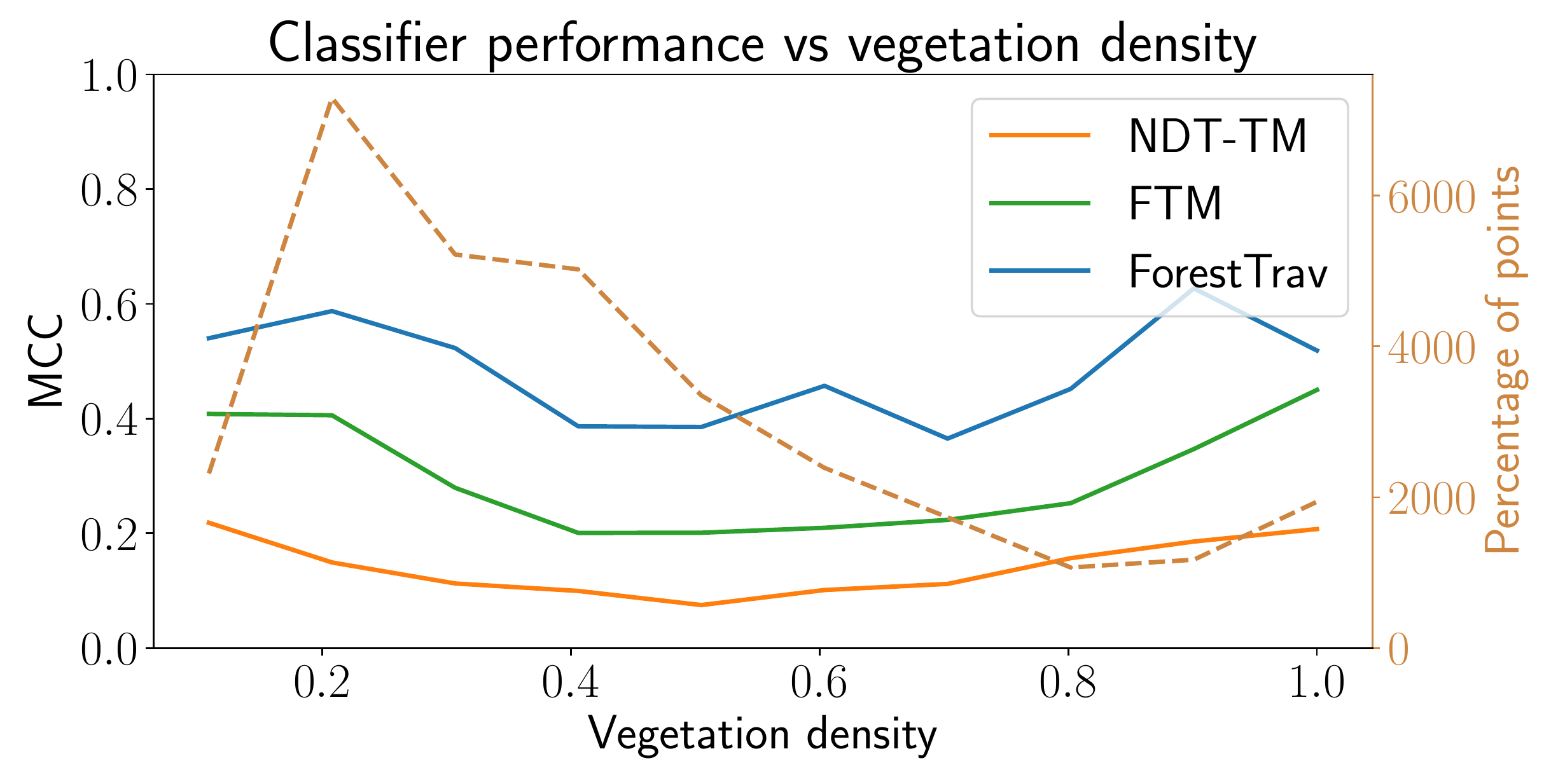}
    \caption
    {The comparison of methods and performance over vegetation density (solid). The number of samples used for each \ac{MCC} score (dashed). }
    \label{fig:mcc_vs_vegd}
\end{figure}
Fig.~\ref{fig:mcc_vs_vegd} shows an overall performance gap between ForesTrav and the other methods for all \ac{vd}. Firstly, we note that ForestTrav shows much higher accuracy than the SOTA methods over the full range of \ac{vd} values. Secondly, all three methods show a trend where performance is higher at low and high \ac{vd}, with reduced performance for the intermediate values of \ac{vd}.  The high performance for the extreme cases of \ac{vd} could be due to seemingly easier cases, where a column contains either little vegetation above the ground (lowest \ac{vd}) or fully contains a large dense obstacle, such as a tree trunk (highest \ac{vd}).  The plot suggests that intermediate \ac{vd} values are the hardest cases to assess accurately, which intuition would also suggest. 
The dashed brown histogram (note the distinct scale on the vertical axis on the right of the graph) indicates the points used per \ac{vd} bin. This indicates that most of the data lies within 0.2 and 0.6 \ac{vd}, even though points above \qty{1}{\m} were included. This shows the need to increase performance in those areas in particular, as that is where the methods struggle most.

\subsubsection{Comparison against SOTA 2.5D methods}
By comparing our approach with two 2.5D geometric elevation map methods, we aim to highlight the challenges and constraints of 2.5D methods in our target environment. The first method, CSIRO-TE, creates a 2.5D elevation map from the same 3D probabilistic voxel representation we use but does not leverage any other salient features. Traversability is assessed using geometric clues, e.g. slope steepness and step height~\cite{HudTal21}. The second method, Elevation Mapping CuPy (EM-CuPy), relies on a GPU-based implementation for elevation mapping. This approach uses a learning-based model with different receptive fields~\cite{miki2022elevation}. \review{We use the same data sources for all methods and tune each approach as best as possible for the current environment.} Both SOTA methods have been successfully demonstrated in long-term navigation during the DARPA SubT challenge~\cite{chung2023into} and used by the top two competitors. 
For a fair comparison, we compressed our 3D maps containing the traversability estimate of our method to 2D grid maps at the same resolution (\qty{0.1}{\m}). Each grid cell is estimated to be either traversable or non-traversable and corresponds to a map column. For each column of the voxel map, we designated the lowest point as the ground voxel and considered all traversability predictions of the set of voxels in the column from, and including, the ground up to 1m. We used a conservative approach, where the corresponding grid cell was considered traversable only if all voxels in the set were traversable. 

\begin{figure}[t]
    \centering
    \includegraphics[width=\columnwidth]{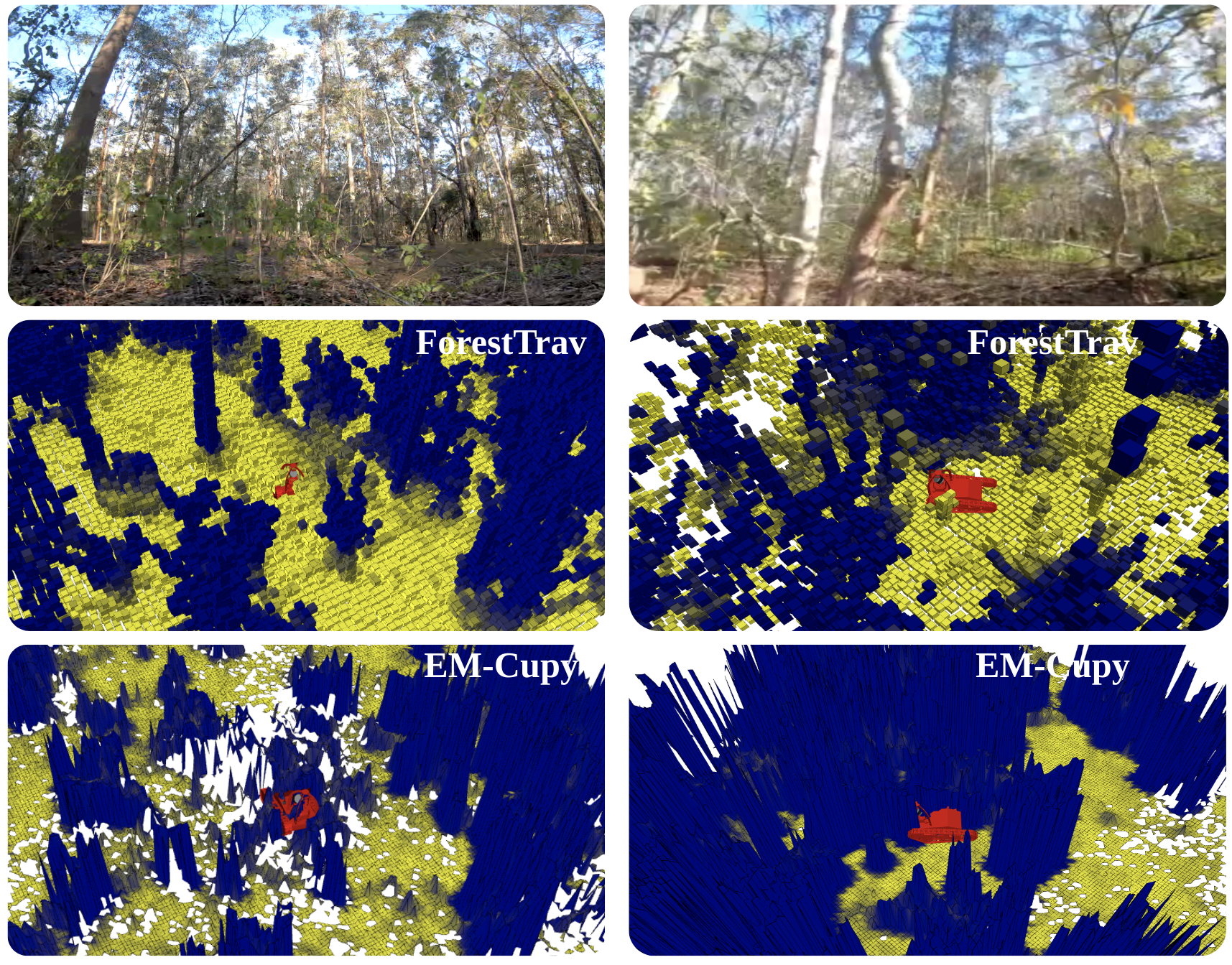}
    \caption{Qualitative comparison of 2.5D representation to our method. The robot model is shown in red. The color indicates TE, yellow (traversable), and blue (non-traversable). Each column shows a different scene. The top row shows the first-person camera view from the robot in the scene, the middle row ForestTrav, and the bottom row EM-CuPy. Left: Area with trees and many thin, small tree stems (traversable) small spaces. ForestTrav correctly assesses them. EM-CuPy wrongly assesses them as non-traversable, and high height variation (spikes) due to many different height measurements. Right: Overhanging branches at different heights.  ForestTrav can represent the environment in full 3D and assess it correctly. EM-CuPy closes off the path due to overhanging obstacles due to limitations of 2.5D representations  }
    \label{fig:2d_como_qulaitative}
\end{figure}

\begin{table}[t]
    \centering
    \caption{Comparison against 2.5D representations}
        \begin{tabular}{lcc}
        \toprule
            Method name &    MCC  & F1-Score    \\ \hline
            EM-CuPy~\cite{miki2022elevation}          &    $ 0.19$ & $0.39$ \\ 
            CSIRO-TE~\cite{HudTal21}                       &   $ 0.14$ & $ 0.38 $\\
            \textbf{ForestTrav (Ours)}                  &    $\bm{0.53}$ & $\bm{ 0.74 }$  \\
        \end{tabular}
    \label{tab:2d_costmap_comparision}
\end{table}
Our 2D cost map computed based on ForestTrav shows the highest performance of all the methods in Table~\ref{tab:2d_costmap_comparision}, indicating the benefits of estimating TE in full 3D before compressing it into 2D. CSIRO-TE maintains a similar voxel representation at the same resolution as our method, but cannot assess the environment accurately. Alternatively, EM-CuPy uses a high-resolution 2.5D map but is still ill-suited for environments with overhanging elements and high clutter, as illustrated in Fig.~\ref{fig:2d_como_qulaitative}. It also fails to deal with grass elements and small bushes since they are assumed to be rigid obstacles. 2.5D geometric representations like this are commonly used as a backbone for SOTA appearance-based or mixed TE estimation methods~\cite{frey2023fast}.

\subsection{Feature ablation}
An ablation study was performed to understand the benefits or limitations of using distribution values directly compared to classical feature calculations. Six different cases are examined; $F_{OCC}$, $F_{RGBO}$, $F_{NDT-TM}$, $F_{FTM}$, and two versions of our method using the distributions directly, $F_{ForestTrav}$ and,  $F_{ForestTrav+RGB}$. The first feature set contains only the occupancy features $F_{OCC} = [N_{OCC}, l_{OCC}]$ from \mbox{\ac{NDT-OM}}~\cite{saarinen2013normal}.  $F_{RGBO}$ contains the occupancy features and additional mean estimation of RGB color channels; $F_{RGB} = [\mu_{R}, \mu_{G}, \mu_{B}]$. $F_{NDT-TM}$ and $F_{FTM}$ are the feature sets used in the respective methods~\cite{ahtiainen2017normal, ruetz2022FTM}. Lastly,  $F_{ForestTrav+RGB}$ is the presented lidar-only method augmented with RGB color information per voxel. The same training and testing procedure is used in section~\ref{res:comparsion_sota}.
\begin{table}[t]
    \centering
    \caption
    {Ablation study for different feature sets used}
    \resizebox{\columnwidth}{!}{
        \begin{tabular}{llcc}
        \toprule
    Feature set& $vs$ [m] &   MCC $ (\mu \pm \sigma)$  & F1-Score $ (\mu \pm \sigma)$  \\
    \hline
    $F_{OCC}$                   & 0.1 &	$ 0.56 \pm 0.06$ & $ 0.78 \pm 0.02$ \\
    $F_{RGBO}$ 	                & 0.1 &	$ 0.55 \pm 0.05$ & $ 0.78 \pm 0.02$ \\
    $F_{NDT-TM}$	            & 0.1 & $ 0.60 \pm 0.06$	 & $ 0.79 \pm 0.04 $\\
    $F_{FTM}$		            & 0.1 & $ 0.56 \pm 0.05$	 & $ 0.78 \pm 0.03 $\\
    $\bm{F_{ForestTrav}}$		& \textbf{0.1} & $ \bm{0.62 \pm  0.03} $ & $\bm{0.82 \pm  0.04}$  \\
    $F_{ForestTrav+RGB}$        & 0.1 & $ 0.61 \pm 0.04 $	& $ 0.81 \pm 0.02 $ \\ 
    \hline
    $F_{OCC}$   	         & 0.2 & 	$ 0.42 \pm	0.03 $ & $	0.70 \pm 0.01 $ \\
    $F_{RGBO}$	              & 0.2 &	$ 0.41 \pm	0.02 $ & $	0.69 \pm 0.01 $ \\
    $\bm{F_{NDT-TM}}$	       & \textbf{0.2} & 	$ \bm{0.56 \pm	0.06 }$ & $ \bm{ 0.77 \pm 0.03 }$ \\
    $F_{FTM}$	                & 0.2 &	$ 0.47 \pm	0.06 $ & $	0.7 \pm 0.02 $ \\
    $F_{ForestTrav}$	        & 0.2 &	$ 0.49 \pm	0.04 $ & $	0.73 \pm 0.02 $ \\
    $F_{ForestTrav+RGB}$        & 0.2 &	$ 0.47 \pm	0.06 $ & 	$0.72 \pm 0.04 $ \\
        \bottomrule
        \end{tabular}
    }
    \label{tab:ablation}
\end{table}
The findings in Table~\ref{tab:ablation} indicate that our method's performance is comparable whether it is using RGB information or the NDT-TM feature set. While the pure NDT-OM occupancy-based variant shows impressive performance, there is a clear indication that additional salient features introduce benefits for both resolutions. It highlights the SCNN's ability to utilize context at \qty{0.1}{\m} resolution and notes that it diminishes at a lower (\qty{0.2}{\m}) resolution. The other variations perform comparably at this lower resolution, except NDT-TM, which outperforms all others significantly. 

\subsection{Estimator Robustness to Map Quality}
\label{exp:siq}
This experiment demonstrates a novel method for evaluating the robustness of a \ac{TE} model with regard to the map quality, as described in Sec.~\ref{method:data_set_generation}. It aims to quantify the accuracy of the TE performance of an \ac{AGV} as it explores a new environment when the map quality is initially low and increases with continuous observation of the environment. In addition, this examines if there is a shift in distribution between the training data (post-processed data) and data acquired in real time. \review{To simulate this process, we extracted voxel maps at discrete time intervals along the robot trajectory. We compared the estimator's prediction of each time-dependent map $M(t)$ against the post-processed ground-truth map $M_{Post}$ for an unobserved test scene (\#9).} The results are comparable to Table~\ref{results:tab_method_comparison} since the same test scene was used.

\begin{figure}[]
    \includegraphics[width=\columnwidth ]{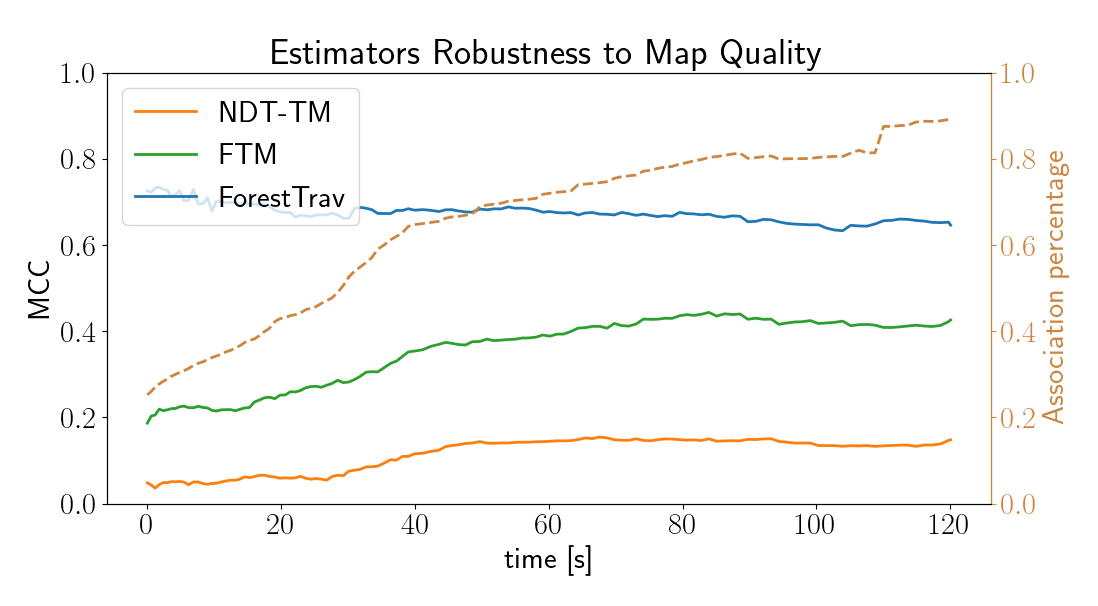}
    \caption{Classification performance of different methods for different times $t$ in an unobserved environment. \review{The brown dashed line shows the association percentage between the temporally evolving probabilistic map $M(t)$ and post-processed ground-truth map $M_{Post}$, a proxy for the amount of the scene explored. All methods were tuned to the best of our capabilities.}}
    \label{fig:results_temporal_performance}
\end{figure}

Fig.~\ref{fig:results_temporal_performance} shows the performance of the TE estimators under this setting. The solid lines illustrate the performance of the compared volumetric approaches over time. The brown dashed line shows the association of \review{the probabilistic map $M(t)$ to the final ground truth map $ M_{Post}$}. It approximates how much of the scene has been observed at least once; the percentage grows continuously until near full coverage. As indicated in Sec.~\ref{method:problem_description}, the differences can be explained by the odometry and SLAM trajectory alignment. 
Our method shows high performance throughout the full evaluation. Performance starts slightly higher than 0.7 \ac{MCC} and slightly decreases to a near-steady state after \qty{\sim 30}{\second}. This indicates a combined reliance on the context and features for accurate TE. A similar MCC score to Table~\ref{res:comparsion_sota} indicates that our method is robust for deployment. The \ac{FTM} method starts below 0.2 \ac{MCC} and increases until it reaches 0.4 \ac{MCC} after \qty{\sim 60}{\second}, subsequently remaining in a steady state with comparable performance to the previous experiment. This indicates that \ac{FTM} is less robust to the map quality but can reach the same performance as in Table~\ref{results:tab_method_comparison}, given sufficient information and time. This makes this method more suitable for offline processing than online deployment, however, it is still much less accurate than our proposed method. 
Similarly, \ac{NDT-TM} performance increases continuously until it reaches a steady state, though it never reaches the same performance as in Table~\ref{results:tab_method_comparison}. This shows that its performance degrades during deployment when exploring novel environments, possible reasons could be overfitting, or a distribution shift between to training data and data acquired during deployment. 
ForestTrav is clearly more robust to map quality than other SOTA methods. This evaluation also offers a practical way to characterize and gauge the suitability of a TE method for exploration. In an exploration task, one would expect to push into new environments continuously. Hence, the strong performance of ForestTrav earlier in the trajectory is significant.

\subsection{Qualitative Evaluation}
\label{sec:qualative_eval}
\begin{figure*}[t]
    \centering
    \includegraphics[width=1.0\textwidth]{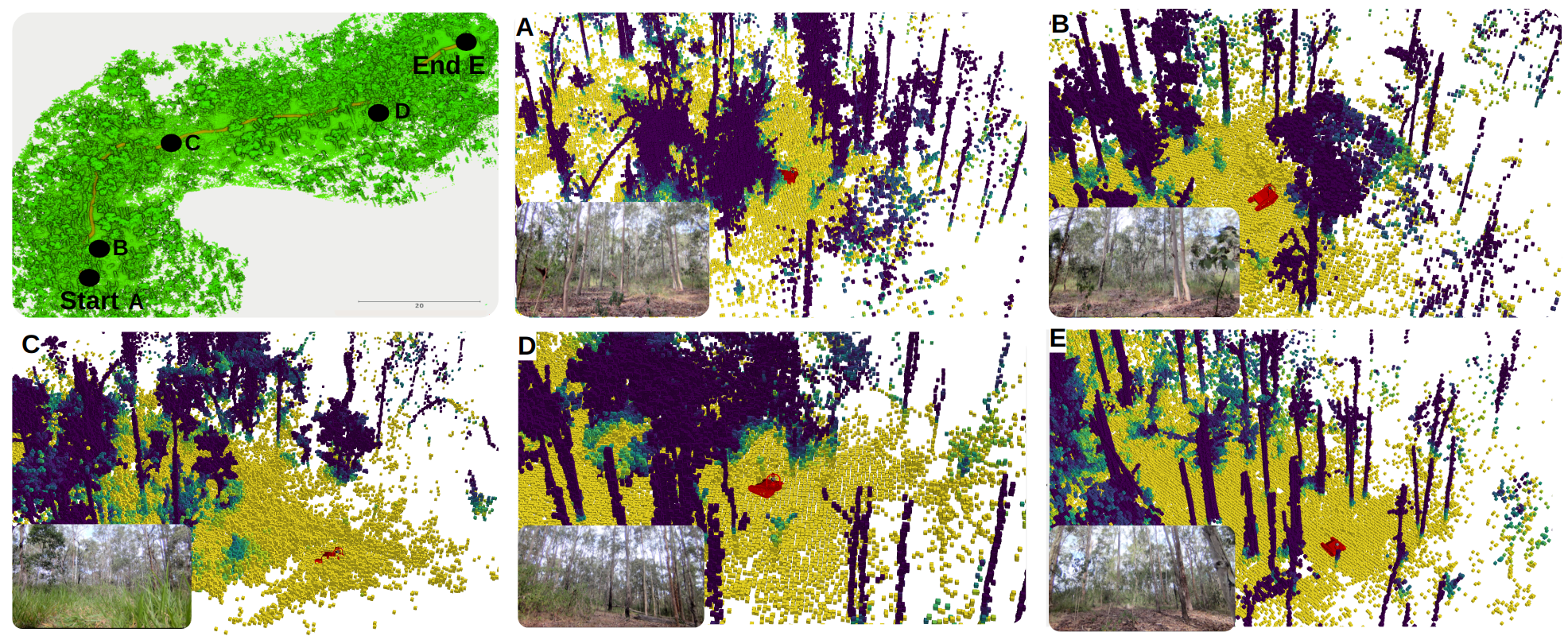}
    \caption{Qualitative examples of online TE of a trajectory in a novel, unknown environment. The top left image shows a bird's eye view of the trajectory with letters indicating the location of the other scenes. Scenes A) - E) show the robot model (red) with the robot's current camera view (bottom left) and the voxel-based traversability probability results, colored as in previous scenes. The images were captured from ForestTrav on real-world data in real time. For this experiment, a different sensor-pack, with the same sensor configuration but double the spin-rate was used, showing generalization to different instances of the same sensor. }
    \label{fig:qualitative_examples}
\end{figure*}
\review{Fig.~\ref{fig:qualitative_examples} shows the TE classification for an exploration setting where the robot drives from A) to E) remotely controlled by an operator and encounters different scenarios in a new, previously unseen environment. The examples are shown for the online inference of our method on raw, real-world sensory data. } A wider and heavier platform, but with similar height and similar sensor configuration (double the spin rate) was used. Image A shows the TE probability \qty{20}{\second} after the mission starts and B approximately \qty{10}{\second} later. From A) different foliage, green elements can be seen, as well as thin stems littering the environment. From the robot's point of view, many parts of the environment are initially obstructed but can then later be observed if the robot moves slightly forward. Single viewpoints can be limiting, with regular occlusions, showing the need for a probabilistic method to fuse sensor readings. In image C, the robot was driving through tall grass. It correctly assessed the vegetation as traversable but had little information about the ground in front of it, as the grass obstructed almost all the environment except the patch near the robot. D) and E) show scenes further into the forest where dense foliage is less prominent, but thin and tall small tree stems are frequent and can be challenging to assess. Overall, our method assessed the environment accurately and qualitatively sensibly. The presented scenes aim to show additional examples in addition to the accompanying video. \review{We note clutter and occlusions are frequent problems, making this a challenging unstructured environment and a probabilistic map helps to alleviate this.}

%% file: 05_discussion_conclusion.tex
\section{Discussion}
ForestTrav demonstrates a novel, real-time method capable of accurate \ac{TE} in densely-vegetated environments. It relies solely on a salient 3D probabilistic representation and leverages contextual information in a principled manner using \ac{SCNN}, showing significantly increased \ac{TE} accuracy compared to the SOTA. We also demonstrated that this method generalizes well, showing qualitatively comparable performance over data sets gathered in these environments.
The feature ablation study suggests that using the distribution values directly avoids the need for additional computational effort for feature calculations, particularly the expensive eigenvalue decomposition. Additionally, we note the performance of the pure NDT-OM, using only occupancy distribution as features ($F_{OCC}$), is quite accurate, but less than the lidar-only method. These findings indicate that the \ac{SCNN} can leverage contextual information from the environment to make accurate decisions but requires sufficient contextual information and resolution. \ac{TE} using occupancy could be sufficient for many environments only containing small amounts of vegetation, making it easy to implement and deploy. Our proposed method is more suitable for higher vegetated environments, whilst the low computational requirements (model size \qty{\sim 17}{\mega\byte}), training time (below 4 minutes), and the low amount of training data still would allow it to be broadly applicable. 

The method shows limitations in cases where the robot moves too fast to gather enough information, the vegetation is too dense for lidar to penetrate or fails to discriminate large bushes or bramble. The method may be over-confident on the map's borders in its traversability prediction, which is common for deep learning methods. \review{Further, our method currently cannot interpolate between missing data or perform any direct form of scene completion, possibly preventing access to an area if it cannot be sufficiently observed. This occurs due to occlusions from vegetation and is a major challenge in vegetated environments. Probabilistic maps mitigate this to a large degree and are superior to single viewpoint assessment due to the ability to fuse previously recorded data from multiple viewpoints. However, the approach will still fail if the density is too high.} Since our method can maintain a large local representation of the environment, it can mitigate some of these issues by re-planning and finding less risky paths and better-observed areas. 

The comparison against SOTA 2.5D methods places our method in the context of recent work deployed by two of the most successful teams in the DARPA SubT Challenge~\cite{chung2023into, HudTal21}. The comparison against CSIRO-TE, using a similar voxel representation, shows the need for salient features and an appropriate estimator to achieve sufficiently accurate performance for real-world deployment in these challenging conditions. On the other hand, the comparison against~\cite{miki2022elevation} highlights the issues that occur even with high resolution (\qty{4}{\cm}) 2.5D maps, even when leveraging SOTA learning methods. Further, SOTA appearance methods commonly fuse visual appearance estimates with these 2.5D representations, e.g.~\cite{frey2023fast}. The data in our target environment suggests these 2.5D representations are insufficient for application in our proposed environments, regardless of the accuracy of the image-based modules. 

Color and semantic information have been reported as powerful features for classification~\cite{wellhausen2019where, kahn2020badgr}. We found that including color features generally shows performance comparable to our lidar-only approach, but can degrade classification performance in some cases, as similarly reported by Bradly et al.~\cite{bradley2015scene}. The lidar-color fusion used was extensively tested and deployed in~\cite{HudTal21}, but environmental differences resulted in increased error. Potential sources are frequent camera occlusions, color distortion due to changing lighting conditions, or voxels containing elements of different colours. The difficulties in these environments require novel, robust solutions to leverage appearance information within a high-resolution voxel representation in a principled manner, beyond the scope of this work. 

The novel analysis of the estimator's robustness to map quality provides an additional introspective tool. The combination of features from different representations is typical for learning-based TE methods. Each voxel, and also each of the statistical distributions (or representation) within a voxel, is also assumed to be independent for computational tractability. Hence, quantifying joint uncertainty or entropy of a mixture of the features/statistics of a single voxel is often infeasible due to these assumptions. Therefore, the presented evaluation of an estimator's robustness is a practical evaluation to assess the TE's performance for methods actively exploring previously unseen areas. 

\section{Conclusion}
In this paper we proposed ForestTrav: a real-time capable learning-based \ac{TE} method effective in challenging dense and cluttered vegetated environments. We showed extensive quantitative and qualitative evaluations that indicate that ForestTrav significantly outperforms SOTA methods. Our approach leverages the environmental context of the \ac{SCNN} and salient features of a 3D probabilistic voxel representation to generate an online capable system. In addition, it is trained solely on real-world data and is cost-effective in terms of training data and time. Through a comprehensive evaluation, we demonstrated the performance and generalization capabilities of our \ac{TE} method, including a novel analysis to assess the \ac{TE} in response to the map quality evolving as the robot explores/moves through and collects more sensor data. Further, we demonstrated that a pure ``geometric method'' is capable of accurately assessing pliable vegetation -- a capability that has not previously been shown in the literature.  

\review{ Current limitations around non- or partially-observed areas could be addressed by considering scene completion. Further investigation into the uncertainty quantification would allow assessment of which areas of the map are poorly observed and which elements would be out-of-distribution of the training data. This combination could enable targeted active learning, where the robot purposely interacts with uncertain areas in the map, encouraging learning in new environments and in areas with previously unseen samples. This would allow the agent to adapt to a novel environment online and in real-time, allowing for fast adaptation and avoiding costly labeling of data for novel environments.}